\documentclass{article}
\long\def\cut#1{}
\usepackage{microtype}
\usepackage{graphicx}
\usepackage{booktabs} 
\usepackage{multibib}
\usepackage{amsmath}
\usepackage[table,xcdraw]{xcolor}
\usepackage{bm}
\usepackage{grffile}
\usepackage{subcaption}
\usepackage{floatrow}
\usepackage[font=footnotesize,figurewithin=none]{caption}
\usepackage{amssymb}
\usepackage{hyperref}
\usepackage{dsfont}
\usepackage[noend]{algpseudocode}

\newcommand{\eg}{\textit{e.\,g., }}
\newcommand{\ie}{\textit{i.\,e., }}

\newcommand{\boldtheta}{\boldsymbol{\theta}}

\newcommand{\boldphit}{\boldsymbol{\phi}^{(t)}}

\newcommand{\boldthetazero}{\boldsymbol{\theta}^{(0)}}
\newcommand{\boldphi}{\boldsymbol{\phi}}
\newcommand{\boldthetat}{\boldsymbol{\theta}^{(t)}}
\newcommand{\xbf}{\textbf{x}}
\newcommand{\zbf}{\textbf{z}}

\newcommand{\ourmethod}{{\it CbAS}}
\newcommand{\dbas}{{\it DbAS}}
\newcommand{\rwr}{{\it RWR}}
\newcommand{\fbvae}{{\it FB-VAE}}
\newcommand{\amvae}{{\it AM-VAE}}
\newcommand{\cempi}{{\it CEM-PI}}
\newcommand{\gb}{{\it GB}}
\newcommand{\gbno}{{\it GB-NO}}
\newcommand{\initset}{\{\xbf_\text{init}\}}
\usepackage{todonotes}
\DeclareMathOperator*{\argmax}{argmax}
\DeclareMathOperator*{\argmin}{argmin}

\newcites{supp}{Supplementary References}

\usepackage[accepted]{icml2019}

\icmltitlerunning{Conditioning by adaptive sampling}

\begin{document}

\twocolumn[
\icmltitle{Conditioning by adaptive sampling for robust design}




\begin{icmlauthorlist}
\icmlauthor{David H. Brookes}{bp}
\icmlauthor{Hahnbeom Park}{uw1,uw2}
\icmlauthor{Jennifer Listgarten}{cs}
\end{icmlauthorlist}

\icmlaffiliation{bp}{Biophysics Graduate Group, UC Berkeley, CA}
\icmlaffiliation{uw1}{Department of Biochemistry, University of Washington, Seattle, WA}
\icmlaffiliation{uw2}{Institute for Protein Design, University of Washington, Seattle, WA}
\icmlaffiliation{cs}{EECS Department, UC Berkeley, CA}

\icmlcorrespondingauthor{David Brookes, Jennifer Listgarten}{\{david.brookes, jennl\}@berkeley.edu}

\vskip 0.3in
]



\printAffiliationsAndNotice{} 



\begin{abstract}
We present a method for design problems wherein the goal is to maximize or specify the value of one or more properties of interest (\eg maximizing the fluorescence of a protein). We assume access to black box, stochastic ``oracle" predictive functions, each of which maps from design space to a distribution over properties of interest. Because many state-of-the-art predictive models are known to suffer from pathologies, especially for data far from the training distribution, the design problem is different from directly optimizing the oracles. Herein, we propose a method to solve this problem that uses model-based adaptive sampling to estimate a distribution over the design space, conditioned on the desired properties.  
\end{abstract}

\section{Predictive-model based design}

The design of molecules and proteins to achieve desired properties, such as binding affinity to a target, has a long history in chemistry and bioengineering. Historically, design has been performed through a time-consuming and costly iterative experimental process, often dependent on domain experts using a combination of intuition and trial and error. For example, a popular technique is {\it Directed Evolution}~\cite{Chen1991}, where first random variations in a parent population of proteins are induced; next, for each of these variants, the property of interest or a proxy to it, is measured; finally, the procedure is repeated on the top-performing variants and the process is iterated. This process is dependent on expensive and time-consuming laboratory measurements. Moreover, it is performing a hybrid greedy-random (uniform) walk through the protein design space---a strategy that is not likely to be particularly efficient in finding the best protein.

Advances in biotechnology, chemistry and machine learning allow for the possibility to improve such design cycles with computational approaches (\eg \citet{Schneider1994,Schneider1998, Gomez-Bombarelli2018,Killoran2017, Brookes2018, Gupta2019, Yang2018}). Where exactly should computational methods be used in such a setting?
Within the aforementioned {\it Directed Evolution}, the obvious place to employ machine learning is to replace the laboratory-based property measurements with a property `oracle'---a regression model that bypasses costly and time-consuming laboratory experiments. However, a potentially higher impact innovation is to also replace the greedy-random search by a more effective search. In particular, given a probabilistic oracle for a property of interest, one can achieve more effective search by employing state-of-the-art optimization algorithms over the inputs (\eg protein sequence) of the oracle regression model, ideally while accounting for uncertainty in the oracle (\eg \citet{Brookes2018}). 

In the framing of the problem just described, there is an implicit assumption that the regression oracle is well-behaved, in the sense that it is trustworthy not only in and near the regime of inputs where it was trained, but also beyond. However, it is now well-known that many state-of-the-art predictive models suffer from pathological behaviour, especially in regimes far from the training data~\cite{Szegedy2014, NguyenYC15}. Methods that optimize these predictive functions directly may be led astray into areas of state space where the predictions are unreliable and the corresponding suggested sequences are unrealistic (\eg correspond to proteins that will not fold). Therefore, we must modulate the optimization with prior information to avoid this problem. There are two related viewpoints on what this prior information represents: 
either as encoding knowledge about the regions where the oracle is expected to be accurate (\eg by representing information about the distribution of training inputs), or as encoding knowledge about what constitutes a `realistic' input (\eg by representing proteins known to stably fold). Herein we focus on the former viewpoint, and thus assume that we have access to the input distribution of our oracle training data. However, the latter viewpoint is required when such data is not available.  

How then should one use such prior knowledge? Formally, for inputs, $\xbf$ and  property of interest $y$, we should model the joint probability $p(\xbf, y)=p(y|\xbf) p(\xbf)$, and then perform design (\ie obtain one or more sequences with the desired properties) by sampling from the conditional distribution. For example, in the case of maximizing one property oracle, we should sample from $p(\xbf|y \geq y_\text{max})$ to obtain our desired designed sequences. More generally, one conditions on the appropriate desired event $p(\xbf|S)$, where $S$ is the conditioning event). 
To achieve this conditioning, we will assume that we have access to a property oracle, $p(y|\xbf)$, which may be a black box function. We also assume that our prior knowledge has been encoded in $p(\xbf)$. The prior density, $p(\xbf)$, can be modelled by training an appropriate generative model, such as a Variational Auto-Encoder (VAE)~\cite{KingmaW13}, Real NVP~\cite{realNVP}, an HMM~\cite{baum1966} or a Transformer~\cite{Vaswani2017,Rives2019}, on the chosen set of `realistic' examples. Recently some generative models have themselves been shown to exhibit pathologies~\cite{nalisnick2018}, however we have not observed any such phenomenon in our setting.

\cut{
\paragraph{Derivative-free oracle}
A main {\it desideratum} for our solution is that the oracle need not be differentiable. In other words, {\it the oracle need only be a black box that provides an input to output mapping}. Such a constraint arises from the fact that in many scientific domains, excellent predictive models already exist which may not be readily differentiable. Additionally, we may choose to use wet lab experiments themselves as the oracle. Consequently, we seek to avoid any solution that relies on differentiating the oracle; although for completeness, we compare performance to such approaches in our experiments.
}

\paragraph{Derivative-free oracle}
One {\it desideratum} for our approach is that the oracle need not be differentiable. In other words, the oracle need only be a black box that provides an input to output mapping. Such a constraint arises from the fact that in many scientific domains, excellent predictive models already exist which may not be readily differentiable. Additionally, we may choose to use wet lab experiments themselves as the oracle. Consequently, we seek to avoid any solution that relies on differentiating the oracle; although for completeness, we compare performance to such approaches in our experiments.

\section{Related work}

The problem set-up just described has a strong similarity to that of {\it activation-maximization} (AM) with a Generative Adversarial Network (GAN) prior~\cite{Goodfelow2014, Simonyan2014, NguyenDYBC16}, which is typically used to visualize what a neural network has learned, or to generate new images. 
\citet{nguyen2017plug} use approximate Langevin sampling from the conditional $p(\xbf|y=c)$, where $c$ is be the event that $\xbf$ is an image of a particular class, such as a penguin.
There are two main differences between our problem setting and that of AM. The first is that AM is conditioning on a discrete class being achieved, whereas our oracles are typically regression models. The second is that our design space is discrete, whereas in AM it is real-valued. Aside from the fact that in any case AM requires a differentiable oracle, these two differences pose significant challenges. The first difference makes it unclear how to specify the desired event to condition on, since the event may be that involving an unknown maximum. Moreover, even if one knew or could approximate this maximum, such an event would be by definition rare or never-seen, causing the relevant conditional probability to yield vanishing gradients from the start. Second, the issue of back-propagating through to a discrete input space is inherently difficult, although attempts have been made to use annealed relaxations~\cite{Jang2017}. In fact, \citet{Killoran2017} adapt the AM-GAN procedure to side-step these two issues for protein design. Thus in our experiments, we compare to their variation of the AM approach.

\citet{Gomez-Bombarelli2018} tackle a chemistry design problem much like our own. Their approach is to (1) learn a neural-network-supervised variational auto-encoder (VAE) latent space so as to order the latent space by the property of interest, (2) build a (Gaussian Process) GP regression model from the latent space to the supervised property labels, (3) perform gradient-based maximisation of the GP over the latent space, (4) decode the optimal solution point(s) using the VAE decoder. Effectively they are approximately modelling the joint probability $p(\xbf, \zbf, y)=p(y|\zbf)p(\xbf|\zbf)p(\zbf)$, for latent representation $\zbf$, and then finding the argument, $\zbf$, that maximizes $E[p(y|\zbf)]$ using gradient descent. Similarly to AM, this approach does not satisfy our black box oracle {\it desideratum}. This approach in turn has a resemblance to \citet{Engel2018}, wherein the goal is image generation, and a GAN objective is placed on top of a VAE in order to learn latent constraints. Because this latter approach was designed for (real-valued) images and for classification labels, we compare only to \citet{Gomez-Bombarelli2018}.

\citet{Gupta2019} offer a solution to our problem, including the ability to use a non-differentiable oracle. They propose to first train a GAN on the initial set of realistic examples and then iterate the following procedure: (1) create a sample set by generating samples from the GAN, (2) use an oracle regression model to make a point prediction for sample as whether or not a protein achieved some desired property, (3) update the sample set by replacing the oldest $n$ samples with the $n$ samples from step 2 that exceed a user-specified threshold that remains fixed throughout the algorithm (and where $n$ at each iteration is determined by this threshold), (4) retrain the GAN on the updated set of samples from step 3 for one epoch. The intended goal is that as iterations proceed, the GAN will tend to produce samples which better maximize the property. They argue that because they only replace $n$ samples at a time, the shift in the GAN distribution is slow, enabling them to implicitly stay near the training data if not run for too long. Their procedure does not arise from any particular formalism. As such, it is not clear what objective function is being optimized, nor how to choose an appropriate stopping point to balance progress and staying near the original realistic examples.

Our approach, {\it Conditioning by Adaptive Sampling} (\ourmethod), offers several advantages over these methods. Our approach is grounded in a coherent statistical framework, with a clear objective, and explicit use of prior information. It does not require a differentiable oracle, which has the added benefit of side-stepping the need to back-propagate through to discrete inputs, or of needing to anneal an approximate representation of the inputs. Our approach is based on parametric conditional density estimation. As such, it resembles a number of model-based optimization schemes, such as Evolutionary Distribution Algorithm (EDA) and Information Geometric Optimization (IGO) approaches~\cite{CMA-ES, IGO2017}, both of which have shown to have good practical performance on a wide range of optimization problems. We additionally make use of ideas from Cross Entropy Methods (CEM) for estimating rare events, and their optimization counterparts~\cite{Rubinstein1997,Rubinstein1999}, which allows us to robustly condition on rare events, such as maximization events. In the case where one can be certain of a perfectly unbiased oracle, \dbas, which uses no prior information, should suffice; this method in turn is related to some earlier work in protein design (\eg \citet{Schneider1994, Schneider1998}).

Finally, our problem can loosely be seen to be related to a flavor of policy learning in Reinforcement Learning (RL) called Reward Weighted Regression (RWR)~\cite{Peters2007}. In particular, if one ignores the states, then RWR can be viewed as an (Estimation of Distribution Algorithm) EDA~\cite{bengoetxea2001}; as such, without the use of any prior to modulate exploration through the policy space, although at times, RWR imposes a per-iteration constraint of movement by way of a Kullback-Leibler (KL)-divergence term~\cite{Peters2010}---note that this offers no global constraint in the sense of a prior, and could be viewed as a formalization of the retaining of old samples in \citet{Gupta2019}. 

Note that our problem statement is different from that of Bayesian Optimization~\cite{Snoek2012} where the goal at each iteration is to decide where to acquire {\it new} ground truth data labels. We are not acquiring any new labels, but our method could be used at the end of Bayesian Optimization (BO) for pure exploitation. Also, a baseline method we introduce, {\it CEM-PI}, could be used to perform optimization within standard BO.

\begin{figure*}[t]
    \centering
    \begin{subfigure}[t]{0.32\textwidth}
        \centering
        \includegraphics[width=\textwidth]{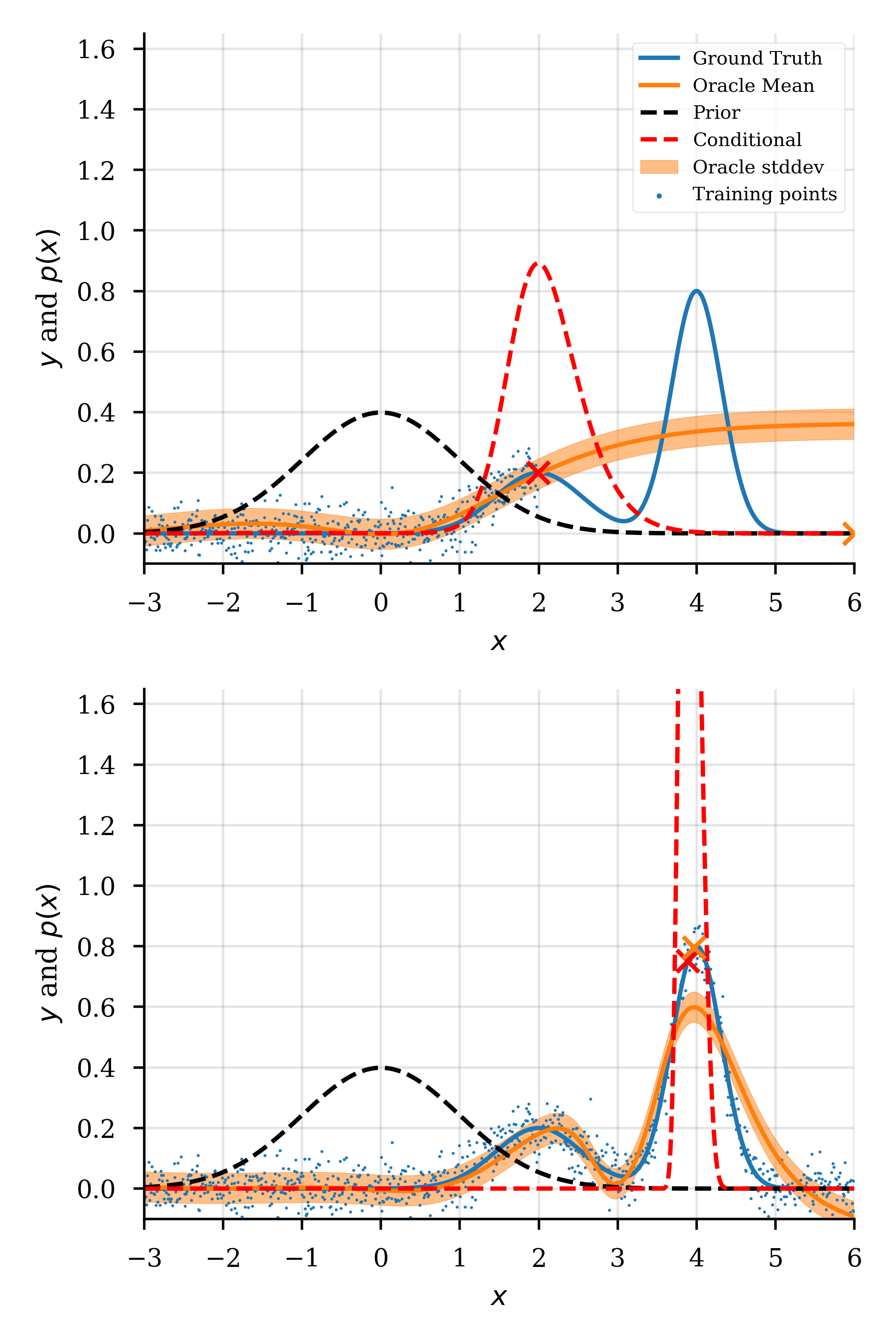}
        \caption{}
    \end{subfigure}
    \begin{subfigure}[t]{0.32\textwidth}
        \centering
        \includegraphics[width=\textwidth]{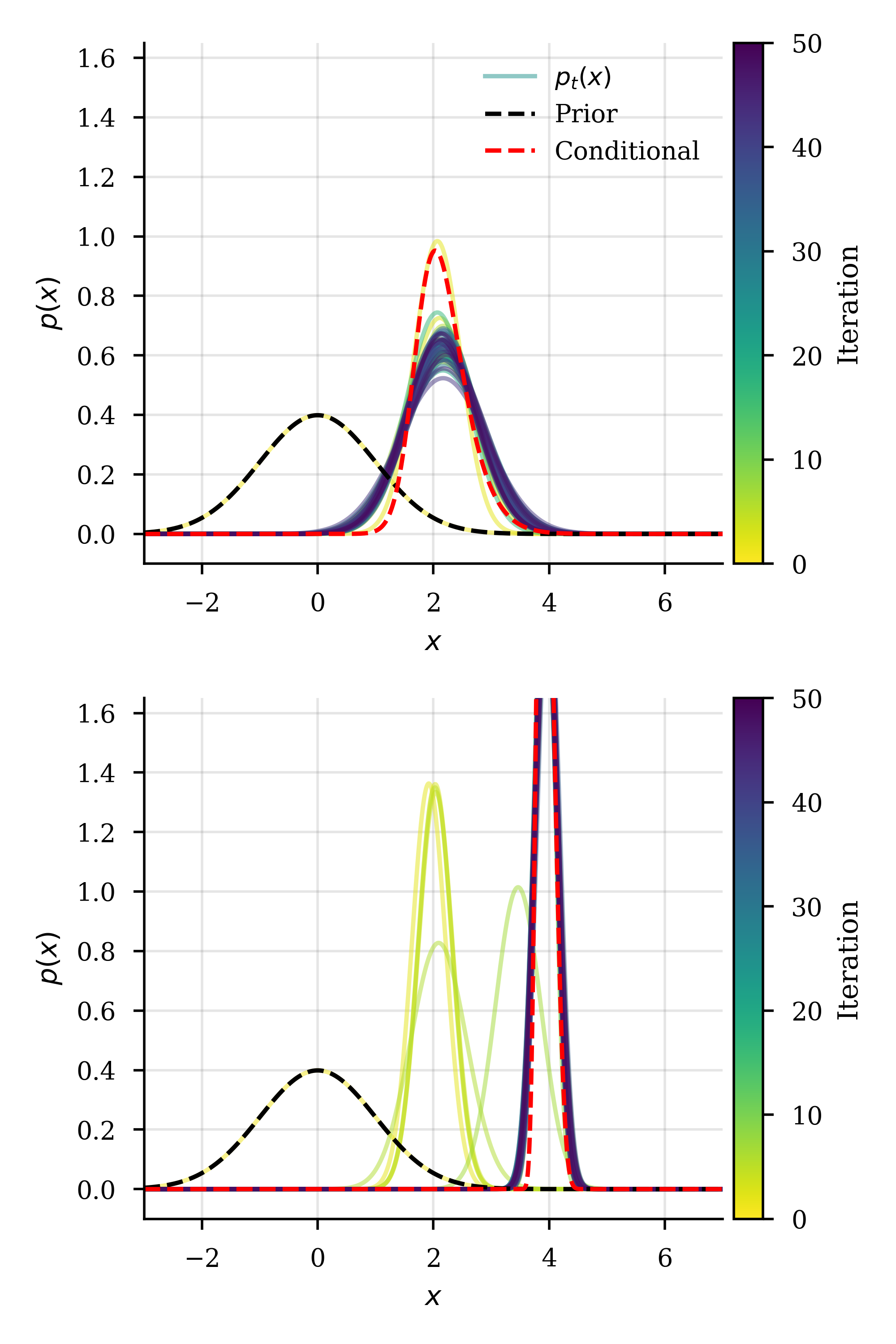}
        \caption{}
    \end{subfigure}
    \begin{subfigure}[t]{0.32\textwidth}
        \centering
        \includegraphics[width=\textwidth]{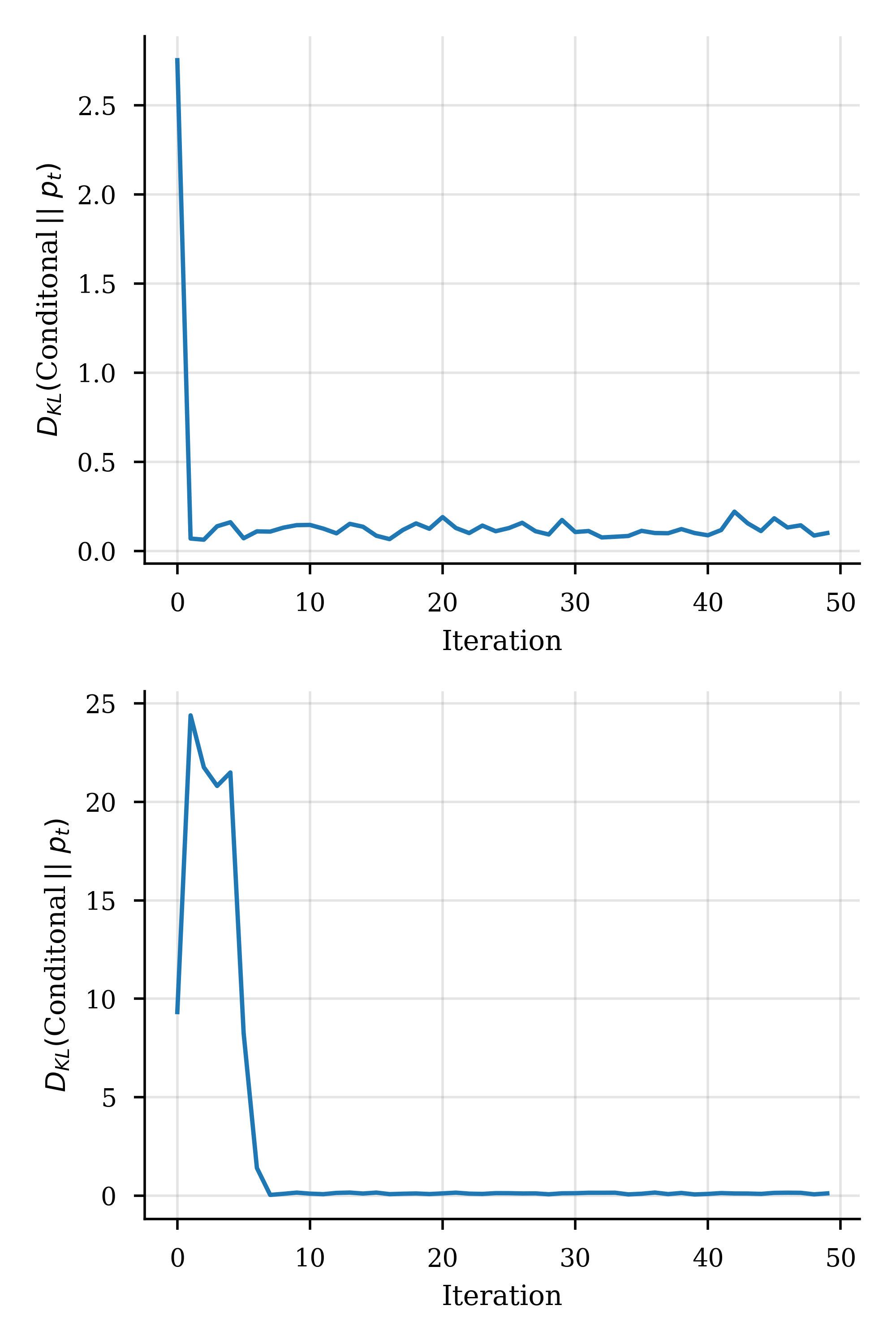}
        \caption{}
    \end{subfigure}
    \vspace{-0.5cm}
    \caption{An illustrative example. (a) Relevant distributions and functions for two oracles (mean and standard deviation shown in orange).
    The oracle in the top plot was given training data corresponding to only half the domain, while the bottom one was given training data covering the whole domain.
    \cut{    trained with different subsets of data drawn from a ground truth function (shown in blue). The prior distribution is shown in black, and t} The prior conditioned on the property that the oracle is greater than its maximum is in red. The value of the ground truth at the mode of the conditional and the maximum of the oracle \cut{(in the plotted range)} are shown as red and orange X's, demonstrating that the mode of the conditional distribution corresponds to a higher value of the ground truth than the maximum of the oracle (b) Evolution (`static animation') of the estimated conditional distribution as \ourmethod~ iterates; the exact distribution is shown in red in panel a. (c) KL divergence between the conditional and search distributions shown in (b), showing that our final approximate conditional is close to the real one.}
    \label{fig:figure1}
\end{figure*}


\section{Methods}\label{methods}

\paragraph{Preamble}

Our problem can be described as follows. We seek to find settings of the $L$-dimensional random vector, $X$ (\eg representative of DNA sequences), that have high probability of satisfying some property {\it desideratum}. For example, we may want to design a protein that is maximally fluorescent (the {\it maximization} problem), or that emits light at a specific wavelength (the {\it specification} problem). We assume that $X$ is discrete, with realizations, $\xbf \in \mathbb{N}^L$, because we are particularly interested in problems of sequence design. However, our method is immediately applicable to $\xbf \in \mathbb{R}^L$, such as images.

We assume that we are given a scalar property predictor ``oracle", $p(y|\xbf)$, which provides a distribution over a property random variable, $Y$ (herein, typically a real-valued property), given a particular input $\xbf$. From this oracle model we will want to compute the probability of various events, $S$, occurring. For maximization design problems, $S$ will be the set of values $y$ such that $y \geq y_\text{max}$ (where \mbox{$y_\text{max} \equiv \max_\textbf{x} \mathbb{E}_{p(y|\textbf{x})}[y]$}).
In specification design problems, $S$ will be the event that the property takes on a particular value, $y=y_\textit{target}$ (strictly speaking, an infinitesimal range around it). In our development, we will also want to consider sets that correspond to less stringent criteria, such as $S$ corresponding to the set of all $y$ for which \mbox{$y \geq \gamma$, with $\gamma \leq y_\text{max}$}.
From our oracle model, we can calculate the conditional probability of these sets---that is, the probability that our property {\it desideratum} is satisfied for a given input---as \mbox{$P(S|\textbf{x}) \equiv P(Y\in S|\textbf{x}) = \int p(y|\textbf{x}) \mathds{1}_{S}(y) \, d y$} (where $\mathds{1}_{S}(y)=1$ when $y \in S$ and 0 otherwise). For the case of thresholding a property value, this turns into a cumulative density evaluation, \cut{\int_{y=\gamma}^{y=\infty} p(y|\textbf{x}) \, d y = }
\mbox{$P(S|\textbf{x}) = p(y \geq \gamma|\textbf{x}) = 1 - CDF(\xbf,\gamma)$}. 

We additionally assume that we have access to a set of realizations of $X$ drawn from some underlying data distribution, $p_d(\xbf)$, which, as discussed above, either represents the training distribution of the oracle inputs, or a distribution of `realistic' examples. Along with these data, we assume we have a class of generative models, $p(\xbf|\boldtheta)$, that can be trained with these samples and can approximate $p_d$ well. We denote by $\boldtheta^{(0)}$ the parameters of this generative model after it has been fit to the data, yielding our prior density, $p(\xbf|\boldthetazero)$. 

Our ultimate aim is to condition this prior on our desired property values, $S$, and sample from the resulting distribution, thereby generating realizations of $\xbf$ that are likely to have low error in the predictive model or are realistic (\ie are drawn from the underlying data distribution); and have high probability of satisfying our {\it desideratum} encoded in $S$. Toward this end, we will assume that the desired conditional can be well-approximated with a sufficiently rich generative model, $q(\xbf|\boldphi)$, which need not be of the same parametric form as the prior.  As our algorithm iterates, the parameter, $\boldphi$, will slowly move toward a value that best approximates our desired conditional density. Our approach has a similar flavor to that of variational inference~\cite{BleiVarTut}, but with a critical difference of needing to handle conditioning on rare events.

Below we outline our approach in the case of maximization of a single property. Details of how to readily generalize this to the specification problem and to more than one property, including a mix of maximization and specification, are in the Supplementary Information.
\vspace{-1 cm}
\paragraph{Our approach}

Our design goal can be formalized as one of estimating the density of our prior  model, conditioned on our set of desired property values, $S$,
\begin{equation}\label{eq:target_conditional}
    p(\xbf|S, \boldthetazero) = \dfrac{P(S|\xbf)p(\xbf|\boldtheta^{(0)})}{P(S|\boldthetazero)},
\end{equation}
where $P(S|\boldthetazero) = \int d\xbf~P(S|\xbf)p(\xbf|\boldtheta^{(0)})$.
In general, there will be no closed-form solution to this conditional density; hence we require a technique with which to approximate it, developed herein. The problem is also harder than it may seem at first because $S$ is in general a rare event (\eg the occurrence of large property value we have never seen).
\cut{
Following the success of model-based optimization schemes for problems of this sort, for example in the reinforcement learning community, we propose a model-based method for estimating \eqref{eq:target_conditional}. Specifically, w
}
To find the parameters of the  search model, $q(\xbf|\boldphi)$, we will minimize the KL divergence between the target conditional, \eqref{eq:target_conditional}, and the search model,
\begin{align}
    \boldphi^* &= \argmin_{\boldphi} D_{KL} \left( p(\xbf|S, \boldthetazero) || q(\xbf|\boldphi) \right) \label{eq:dkl1} \\
    &= \argmin_{\boldphi} -\mathbb{E}_{p(\xbf|S, \boldthetazero)}[\log q(\xbf|\boldphi)] - H_0 \\
    &= \argmax_{\boldphi} \dfrac{1}{P(S|\boldthetazero)}\mathbb{E}_{p(\xbf|\boldthetazero)}[P(S|\xbf)\log q(\xbf|\boldphi)] \\
    &= \argmax_{\boldphi} \mathbb{E}_{p(\xbf|\boldthetazero)}[P(S|\xbf)\log q(\xbf|\boldphi)],
    \label{eq:obj1}
\end{align}

where $H_0 \equiv - \mathbb{E}_{p(\xbf|S, \boldthetazero)}[\log p(\xbf|S, \boldthetazero)]$ is the entropy of the target conditional distribution. Neither $H_0$ nor  $P(S|\boldtheta^{(0)})$ rely on $\boldphi$ and thus drop out of the objective.  

The objective in \eqref{eq:obj1} may seem readily solvable by drawing samples, $\xbf_i$, from the prior, $p(\xbf|\boldthetazero)$, to obtain a Monte Carlo (MC) approximation to the expectation in \eqref{eq:obj1}; this results in a simple weighted maximum likelihood problem. However, for most interesting design problems, the desired set $S$ will be exceedingly rare,\footnote{If not, then the design problem was an easy one for which we did not need specialized methods.} and consequently $P(S|\xbf)$ will be vanishingly small for most $\xbf$ sampled from the prior. Thus in such cases, an MC approximation to \eqref{eq:obj1} will exhibit high variance and require an arbitrarily large number of samples to calculate accurately. 
\cut{
The weighted maximum likelihood estimate resulting from the objective using this approximation will also experience high variance.} 
Any gradient-based approach, such as reparameterization or the log-derivative trick~\cite{KingmaW13, RezendeMW14}, to try to solve \eqref{eq:obj1} directly will suffer from a similar problem.

To overcome this problem of rare events, we draw inspiration from CEM \cite{Rubinstein1997, Rubinstein1999} and \dbas~ \cite{Brookes2018} to propose an iterative, adaptive, importance sampling-based estimation method. 

First we introduce an importance sampling distribution, $r(\xbf )$, and  rewrite the objective function in \eqref{eq:obj1} as
\begin{align}
    &\mathbb{E}_{p(\xbf|\boldthetazero)}[P(S|\xbf)\log q(\xbf|\boldphi)] \label{eq:im_exp1}\\
    =& \mathbb{E}_{r(\xbf)}\left[\dfrac{p(\xbf|\boldthetazero)}{r(\xbf)}P(S|\xbf)\log q(\xbf|\boldphi)\right],
    \label{eq:im_exp2}
\end{align}

to mitigate the problem that the expectation of $P(S|\xbf)$ over our prior in \eqref{eq:obj1} is likely to be vanishingly small. 
Now the question remains of how to find a good proposal distribution, $r(\xbf)$. Rather than finding a single proposal distribution, we will construct a series 
relaxed conditions, $S^{(t)}$, and corresponding
importance sampling distributions, $r^{(t)}(\xbf)$, such that
(a) $E_{r^{(t)}(\xbf)}[P(S^{(t)}|\xbf)]$ is non-vanishing, and (b) \mbox{ $S^{(t)} \supset S^{(t+1)} \supset S$}, for all $t$. The first condition implies that we can draw samples, $\xbf$, from $r^{(t)}(\xbf)$ that have reasonably high values of $P(S^{(t)}|\xbf)$. The second condition ensures that we slowly move toward our desired property condition; that is, it ensures that $S^{(t)}$ approaches $S$ as $t$ grows large.
(In practice, we choose to use the less stringent condition \mbox{ $S^{(t)} \supseteq S^{(t+1)} \supseteq S$}, but the stricter condition can trivially be achieved with a minor change to our algorithm, in which case one should be careful to ensure that the variance of the expectation argument does not grow too large).

The only remaining issue is how to construct these sequences of relaxed events and importance sampling proposal distributions. Next we outline how to do so when the conditioning event is a maximization of a property, 
and leave the specification problem to the Supplementary Information.

Consider the first condition, which is that $E_{r^{(t)}(\xbf)}[P(S^{(t)}|\xbf)]$ is non-vanishing. To achieve this, we could (1) set $S^{(t)}$ to be the relaxed condition that $p(y \geq \gamma^{(0)}| \xbf)$ where $\gamma^{(0)}$ is the $Q^\text{th}$ percentile of property values predicted for those samples used to construct the prior, and (2) set the proposal distribution to the prior, $r^{(t)}(\xbf)=p(\xbf|\boldthetazero)$.
For $Q$ small enough, $E_{q(\xbf|\boldphit)}[P(S^{(t)}|\xbf)]$ will be non-vanishing by definition and condition (a) will be satisfied. Thus, by construction, we can now reasonably perform maximization of the objective in \eqref{eq:im_exp2} instantiated with $S^{(t)}$ because the rare event is no longer rare. In fact, this is how we set the first tuple of the required sequence, $(S^{(0)},r^{(t)})$. After that first iteration, we will then have solved for our approximate conditional, under a relaxed version of our property {\it desideratum} to obtain $q(\xbf|\boldphi^{(0)})$.
Then, at each iteration, we set $r^{(t)}=q(\xbf|\boldphi^{(t-1)})$, and $\gamma^{(t)}$ to the $Q^\text{th}$ percentile of property values predicted from the samples obtained in the $(t-1)^{\text{th}}$ iteration. By the same arguments made for the initial tuple of the sequence, we will have achieved condition (a), and condition (b) can trivially be achieved by disallowing $\gamma^{(t)}$ from decreasing from the previous iteration (\ie, set it to the same value as the previous iteration if necessary).

Altogether now, \eqref{eq:im_exp2} becomes
\begin{align}
     \mathbb{E}_{q(\xbf|\boldphit)}\left[\dfrac{p(\xbf|\boldthetazero)}{q(\xbf|\boldphit)}P(S^{(t)}|\xbf)\log q(\xbf|\boldphi)\right] \label{eq:im_exp3},
\end{align}
which we can approximate using MC with samples drawn from the search model, $\xbf_i^{(t)}\sim q(\xbf|\boldphit)$ for $i=1,...,M$, and where $M$ is the number of samples taken at each iteration,
\begin{align}
     \boldphi^{(t+1)} = \argmax_{\boldphi} \sum_{i=1}^M \dfrac{p(\xbf^{(t)}_i|\boldthetazero)}{q(\xbf^{(t)}_i|\boldphit)}P(S^{(t)}|\xbf_i^{(t)})\log q(\xbf^{(t)}_i|\boldphi). \label{eq:mc_obj}
\end{align}
 At last, we now have a low-variance estimate of our objective (or rather, a relaxed version of it that will get annealed), which can be viewed as a weighted maximum likelihood with weights given by $\dfrac{p(\xbf^{(t)}_i|\boldthetazero)}{q(\xbf^{(t)}_i|\boldphit)}P(S^{(t)}|\xbf_i^{(t)})$. Our objective function can now be optimized using any number of standard techniques for training generative models. (In the Supplementary Information we show how to extend our method to models that can only be trained with variational approximations to the maximum likelihood objective). 

It is clear from \eqref{eq:mc_obj} that the variance of the MC estimate not only relies on $P(S^{(t)}|\xbf)$ being non-vanishing for many of the samples but also that the density ratio, $\frac{p(\xbf|\boldthetazero)}{q(\xbf|\boldphit)}$ is similarly well-behaved. Fortunately, this is enforced by the weighting scheme itself, which encourages the density of the search model to remain close to the prior distribution. This is intuitively satisfying, as it shows that minimizing the KL divergence between the search distribution and the target conditional requires balancing the maximization the probability of $S^{(t)}$ with adherence to the prior distribution.

\paragraph{Extension to intractable latent variable models}

Our final objective in \eqref{eq:mc_obj} requires us to reliably evaluate the densities of the prior and search models for a given input $\xbf$. This is often not possible, particularly for latent variable models where the marginalization over the latent space is intractable. Although one might consider using an Evidence Lower Bound (ELBO)~\cite{BleiVarTut} approximation to the required densities, one can exploit the structure of our objective to exactly derive the needed quantity.
\cut{
where one models the joint distribution of $X$ and a random variable $Z$ with realizations $\zbf \in \mathcal{Z}$. Often $\mathcal{Z} \subseteq \mathbb{R}^d$ for some large $d$ and therefore calculating the density of a particular $\xbf$ requires marginalizing a high-dimensional joint distribution. }
In particular, we can maintain exactness of our objective
for prior and search model densities where one can only calculate the joint densities of the observed and latent variables, namely $p(\xbf, \zbf|\boldthetazero)$ and $q(\xbf, \zbf|\boldthetat)$, which can be achieved only if both model's densities are defined on the same latent variable space, $\mathcal{Z}$.

This extension relies on the fact that an expectation over a marginal distribution is equal to the same expectation over an augmented joint distribution, for instance \mbox{$\mathbb{E}_{p(x)}[f(x)] = \mathbb{E}_{p(x, y)}[f(x)]$}. Starting with \eqref{eq:im_exp1} and using this fact, we arrive at an equivalent objective function,
\begin{align}
    &\mathbb{E}_{p(\xbf|\boldthetazero)}[P(S^{(t)}|\xbf)\log q(\xbf|\boldphi)] \\
    =& \mathbb{E}_{p(\xbf, \zbf|\boldthetazero)}[P(S^{(t)}|\xbf)\log q(\xbf|\boldphi)] \\
    =& \mathbb{E}_{q(\xbf, \zbf|\boldphit)}\left[\dfrac{p(\xbf, \zbf|\boldthetazero)}{q(\xbf, \zbf|\boldphit)}P(S^{(t)}|\xbf)\log q(\xbf|\boldphi)\right]. \label{eq:joint_exp}
\end{align}
This objective can then be optimized in a similar manner to the originally presented case, only now using an MC approximation with joint samples $\xbf^{(t)}_i, \zbf^{(t)}_i \sim q(\xbf, \zbf|\boldphit)$. 
\cut{This then allows to solve a corresponding weighted maximum likelihood problem similar to that of \eqref{eq:mc_obj} that minimizes the $D_{KL}$ between the search and target conditional distributions.}

Note that in the common case where both models are constructed such that they have the same density over $\mathcal{Z}$, $p(\zbf)$, then \eqref{eq:joint_exp} can be further simplified using 
\mbox{$\frac{p(\xbf| \zbf,\boldthetazero)p(\zbf)}{q(\xbf| \zbf,\boldphit)p(\zbf)} = \frac{p(\xbf| \zbf,\boldthetazero)}{q(\xbf| \zbf,\boldphit)}$}.
\paragraph{Practical Considerations}

In practice, we often use the same parametric form for the search model and the prior density. This allows us to simply initialize the search distribution parameters as $\boldphi^{(1)} = \boldthetazero$. 
\cut{Otherwise, however, one could fit $\boldphi^{(1)}$ to the data used to train the prior.}
Additionally, in practice we cache the search model parameters $\boldphi^{(t)}$ and use these to initialize the parameters at the next time step in order to reduce the computational cost of the training procedure at each step.

In Algorithm 1 in the Supplemental Information, we outline our complete procedure when the prior and generative model are both latent variable models of the same parametric form (\eg both a VAE).

\section{Experiments}

We perform two main sets of experiments. In the first, we use  simple, one-dimensional, toy examples to demonstrate some of the main points about our approach. In the second set of experiments, we ground ourselves in a real protein fluorescence data set, conducting extensive simulations to compare methods.

\subsection{An illustrative example}

We first wanted to investigate the properties of our approach with a simple, visual example to understand how the prior influences the solutions found. In particular we wanted to see how an oracle could readily become untrustworthy away from the training data, and how use of the prior might alleviate this. We also wanted to see that even when the oracle is trustworthy, that our approach still yields sensible results. Finally, we wanted to ensure that our approach does indeed approximate the desired conditional distribution satisfactorily for simple cases that we can see. The summary results of this experimentation are shown in Figure \ref{fig:figure1}. 

To run this set of experiments, we first constructed a ground truth property function, comprising the superposition of two unnormalized Gaussian bumps. The goal was to find an input, $x$, that maximizes the ground truth, when only the oracle and a prior are given.

Next we created two different oracles, by giving one access only to data that covered half of the domain, and the other by giving it data that covered the entire domain (including all those in the first data set). These training data were ground truth observations corrupted with zero-mean Gaussian noise with variance of 0.05. Then two oracles were trained, one for each data set, and of the form,
$p(y|x)=\mathcal{N}(\mu(x), \sigma^2)$ where $\mu(x)$ is a two hidden-layer neural network fit to one of the training sets and $\sigma^2$ was set to the mean squared error between the ground truth and $\mu(x)$ on a hold-out set.

\cut{
We then drew two sets of training data from this ground truth, with a small amount of noise added. One set only covered part of the important regions of the ground truth (\ie~ the regions with local minimum and maximum), while the other covered nearly all of them. We then constructed two simple oracles with the form $p(y|x)=\mathcal{N}(\mu(x), \sigma^2)$ where $\mu(x)$ is a two hidden-layer neural network fit to one of the training sets and $\sigma^2$ was fit to the mean squared error between the ground truth and $\mu(x)$ on a hold-out set.
}

The resulting oracles, and the underlying ground truth are shown in Figure \ref{fig:figure1}a. The oracle trained with the smaller data set suffers from a serious pathology---it continues to increase in regions where the ground truth function rapidly decreases to zero. This is exactly the type of pathology that \ourmethod~aims to overcome by estimating the conditional density of the prior rather than directly optimizing the objective. The second oracle does not suffer from as serious a pathology, and serves to show that \ourmethod~can still perform well in the case that the oracle is rather accurate (\ie the prior does not overly constrain the search). As with any Bayesian method with a prior, there may be settings where the prior could lead one astray, and this example is simply meant to convey some intuition. The next set of experiments, grounded in a protein design problem, suggest that the prior we constructed, used within \ourmethod, works well in practice.

We construct our target set $S$ as being the set of values for which $Y$ is greater than the maximum of the oracle's expectation, for $x$ values between minus three and six. In this simple 1D case, we can evaluate the target conditional density very accurately by calculating $P(S|x)p_0(x)$ for many values of $\xbf$ and using numerical quadrature to estimate the normalizing constant. The target conditional is shown in red in \ref{fig:figure1}a. We can see that in both cases, the mode of the conditional lies near a local maximum and the conditional assigns little density to regions where the oracle is highly biased. 

Finally, we test the effectiveness of \ourmethod~in estimating the desired conditional densities (Figure \ref{fig:figure1}b,c). We use a search distribution that is the same parametric form as the prior, that is, a Gaussian distribution and run our method for 50 iterations, with the quantile update parameter, $Q=1$ (meaning that $\gamma^{(t)}$ will be set to the maximum over the sampled mean oracle values at iteration $t$) and $M=100$ samples taken at each iteration. Figure \ref{fig:figure1}b shows the search distributions that result from our scheme at each iteration of the algorithm, overlayed with target conditional distribution. Figure \ref{fig:figure1}c) shows the corresponding KL divergences between the search and target distributions as the method proceeds. We can see qualitatively and quantitatively (in terms of the KL divergence) that in both cases the distributions converge to a close approximation of the target distribution. 

\begin{figure*}[!ht]
    \centering
    \begin{subfigure}[t]{0.49\textwidth}
        \centering
        \includegraphics[width=\textwidth]{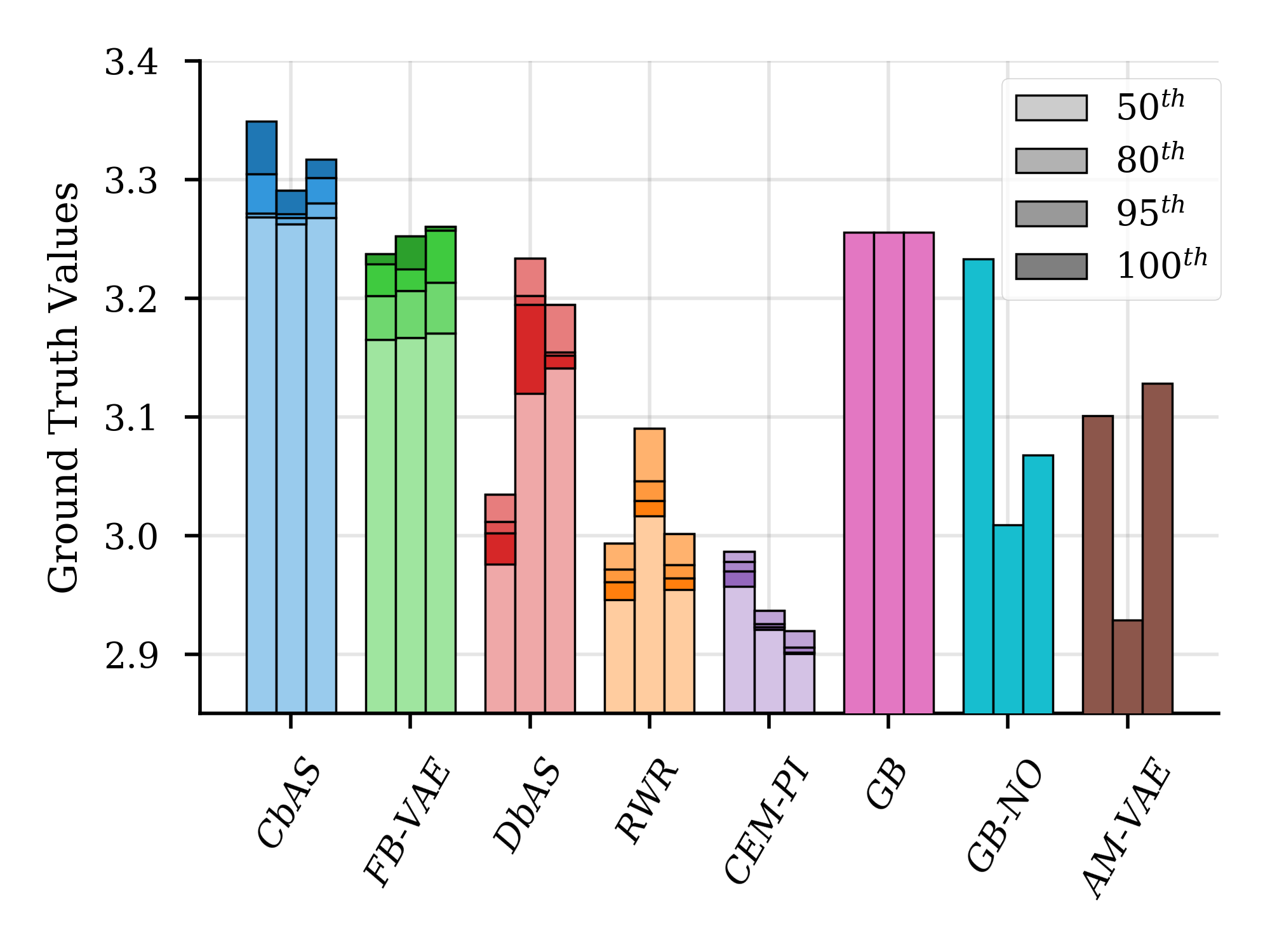}
        \caption{}
    \end{subfigure}
    \begin{subfigure}[t]{0.49\textwidth}
        \centering
        \includegraphics[width=\textwidth]{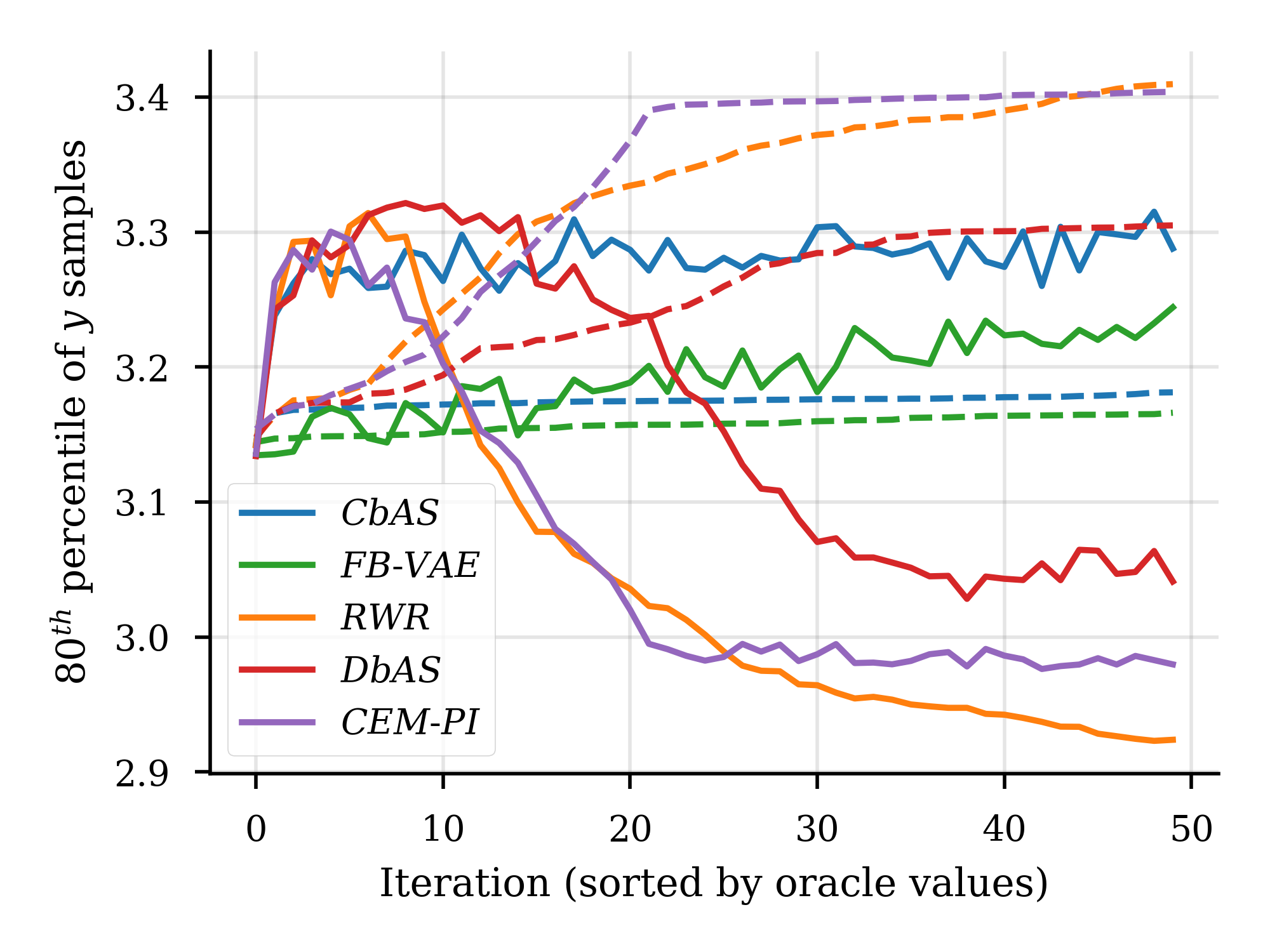}
        \caption{}
    \end{subfigure}
    \vspace{-0.5cm}
    \caption{Design for maximization of protein fluorescence. (a) For sampling-based methods, namely \ourmethod, \fbvae, \dbas, \rwr~ and \cempi, shown are the mean values of the ground truth evaluated at samples coming from different percentiles ($50^\text{th}$, $80^\text{th}$, $95^\text{th}$, and $100^\text{th}$) of oracle predictions over all iterations. The gradient-based methods, \gb,  \gbno~ and \amvae, yield only a single protein at each iteration and converge rapidly; thus only the ground truth value of the final single protein for each of these methods is used. For sampling-based methods, an optimal method has darkest bars at the top, followed by progressively less dark bars, indicating that the method has successfully avoided untrustworthy regions of the space. Methods that have the darkness out of this order are being led astray into untrustworthy oracle regions. The height of a bar shows how well the ground truth fluorescence is maximized, where higher is better. For \ourmethod, the height is highest, and the darkness ordering, correct---unsurprisingly, as avoiding oracle pathologies should help achieve higher ground truth during maximization. This trend of better darkness ordering yielding high property values holds for all sampling based methods. The sets of three bars indicate the three different oracles (each bar was averaged over three random runs). (b) For one representative run in panel a), a trajectory is shown for each method (other runs are in the Supplemental Information). Each point on a dashed lines shows the $80^{th}$ percentile of oracle evaluations of the samples at that iteration. The corresponding point on a solid line shows the mean ground truth value of those same samples. The last point on a curve shows what final protein sequence would be used from each method (\ie that with the highest oracle value seen, as the ground truth would be unknown)---high dashed values and low solid ones are methods that have been led astray into pathological regions of the oracle. }
    \label{fig:figure2}
\end{figure*}

\subsection{Application to protein fluorescence} \label{GFP_section}

Fluorescent proteins are a workhouse of modern molecular biology. Improving them has a long history, but also serves as a good test bed for protein optimization. Thus we here focus on the problem of maximizing protein fluorescence.
In particular, we perform a systematic comparison of our method, \ourmethod, to seven other approaches, described below. We anchored our experiments on a real protein fluorescence data set~\cite{Sarkisyan2016} (see Supplementary Information).
\vspace{-0.25cm}
\paragraph{Methods considered}
We compared our approach to other competing methods, including, for completeness, those which can only work with differentiable oracles. For models that originally used a GAN prior, we instead use a VAE prior so as to make the comparisons meaningful.\footnote{As shown in \citet{Brookes2018}, the GAN and VAE appear to yield roughly similar results in this problem setting.} We compare our method, \ourmethod~ against the following methods: 
(1) \amvae---the activation-maximization method of \citet{Killoran2017}. This method requires a differentiable oracle.
(2) \fbvae---the method of \citet{Gupta2019}. This method does not require a differentiable oracle.
(3) \gbno---the approach described by ~\citet{Gomez-Bombarelli2018}. This method requires a differentiable oracle.
(4) \gb ---the approach implemented by ~\citet{Gomez-Bombarelli2018} which has some additional optimization constraints placed on the latent space that were not reported in the paper but were used in their code. This method requires a differentiable oracle.
(5) \dbas---a method similar to \ourmethod, but which assumes an unbiased oracle and hence has no need for or ability to incorporate a prior on the input design space.
(6) \rwr---Reward Weighted Regression~\cite{Peters2007}, which is similar to \dbas, but without taking into account the oracle uncertainty. This method does not require a differentiable oracle.
(7) \cempi---use of the Cross Entropy Method to maximize the Probability of Improvement~\cite{Snoek2012}, an acquisition function often used in Bayesian Optimization; this approach does not make use of a prior on the input design space. This method does not require a differentiable oracle.
Implementation details for each of these methods can be found in the Supplementary Information.

In these experiments we set the quantile update parameter of \ourmethod~to $Q=1$. However, we show in the Supplementary Information that results are relatively insensitive to the setting of $Q$.
\vspace{-0.25cm}
\paragraph{Simulations}

In order to compare methods it was necessary to first simulate the real data with a known ground truth because evaluating hold out data from a real data set is not feasible in our problem setting. To obtain a ground truth model we trained a GP regression (GPR) model on the protein fluorescence data, with a protein-specific kernel~\cite{Shen2014}, whose feature space was augmented by adding a bias feature and exponentiating to obtain a second order polynomial kernel. Our ground truth is the mean of this GPR model, which, notably, is a different model class than the oracle, as we expect in practice. 

Next, for each protein sequence in the original data set we compute its ground truth fluorescence. Then we take those sequences that lie in the bottom $20^{\text{th}}$ percentile of ground truth fluorescence, choose 5,000 of them at random, and use these to train our oracles using maximum likelihood. We wanted to investigate different kinds of oracles with different properties, with a particular focus on investigating different uncertainty estimates. Specifically, we considered three types of oracle (details are in the Supplementary Information). The first was a single-layer neural network model with homoscedastic, Gaussian noise in its predictions. The second was an ensemble of five of these models, each with heteroscedastic noise, as in~\citet{Lakshmin2017}. The third was the same as the second, but with an ensemble of 20 neural networks. Performance of these models on a test set are shown in the Supplemental Information.

Next we train a standard VAE prior (details are in the Supplementary Information) on those same 5,000 samples. This VAE gets used by \ourmethod~ and \amvae~ (as the prior) and in \fbvae~ (for initialization). Because \gb~ and \gbno~ require training a supervised VAE, this is done separately, but with the same 5,000 samples, and corresponding ground truth values.

To fairly compare the methods we keep the total number of samples considered during each run, $N$, constant. We call this the sequence budget. This sequence budget corresponds to limiting the total number samples drawn from the generative model in \ourmethod, \dbas, \fbvae, \rwr~ and \cempi; and limiting the number of total gradient step updates performed in the \amvae, \gb~ and \gbno~ methods. For the latter class of methods, if the method converged but had not used up its sequence budget, then the method was re-started with a new, random initialization and executed until the sequence budget was exhausted, or a new initialization was needed. For convergence we used the built-in method that comes with \gb~ and \gbno, and for \amvae, convergence is defined as that the maximum value did not improve in 100 iterations. 

For each of the three oracle models, we ran three randomly initialized runs of each method with a sequence budget of 10,000. Figure \ref{fig:figure2}a shows the results from this experiment, averaged over the three separate runs. Methods that have no notion of a prior (\dbas, \rwr, \cempi) are clearly led astray, as they only optimize the oracle, finding themselves in untrustworthy regions of oracle space. This suggests that our simulation settings are in the regime of the illustrative example shown in Figure \ref{fig:figure1}a (top). 

\cut{To further sanity check that these models are being led astray, we also roughly estimated the protein stability~\cite{Park2016} of the final selected sequences for each method, finding that that prior-free models can easily be led into parts of the space that are not realistic for exiting proteins because they are unstable (see Supplementary Information).}



 \section{Discussion}

\cut{Herein we assumed that our prior density, captured by a VAE, was appropriately capturing our knowledge about the domain of interest. However, this task alone is not necessarily an easy one, although pursuing it is beyond the scope of this paper. In terms of the modelling aspect of this, intuitively, if one doubles the amount of realistic data available to the modelling of the prior, then we expect it to be more sure of itself. To properly achieve this one may wish to explore how to capture uncertainty of the prior, generative model parameters themselves, for example in the Bayesian sense. }

Our main contributions have been the following. (1) We introduced a new method that enables one to solve design problems in discrete space, and with a potentially non-differentiable oracle model, (2) we introduced a new way to perform approximate, conditional density modelling for rich classes of models, and importantly, in the presence of rare events, (3) we showed how to leverage the structure of latent variable models to achieve exact importance sampling in the presence of models whose density cannot be computed exactly. Finally, we showed that compared to other alternative solutions to this problem, even those which require the oracle to be differential, our approach yields competitive results.

\cut{In future work we plan to further understand the interplay between accuracy of extrapolation and uncertainty estimation, and in particular in the context of the design problem.}


\section*{Acknowledgements} 

The authors thank Sergey Levine, Kevin Murphy, Petros Koumoutsakos, Gisbert Schneider, David Duvenaud, Yisong Yue and Ben Recht for helpful discussion and pointers; and Benjamin Sanchez-Lengeling for providing python scripts to reproduce the method in \citet{Gomez-Bombarelli2018}.

This research used resources of the National Energy Research
Scientific Computing Center, a DOE Office of Science User Facility
supported by the Office of Science of the U.S. Department of Energy
under Contract No. DE-AC02-05CH11231

\bibliography{main}
\bibliographystyle{icml2019}

\clearpage

\setcounter{figure}{0}
\setcounter{section}{0}
\setcounter{equation}{0}
\setcounter{table}{0}
\renewcommand{\thefigure}{S\arabic{figure}}
\renewcommand{\thesection}{S\arabic{section}}
\renewcommand{\theequation}{S\arabic{equation}}
\renewcommand{\thetable}{S\arabic{table}}

\onecolumn
\icmltitle{Supplementary Information: Conditioning by adaptive sampling for robust design}
\icmltitlerunning{Supplementary Information: Conditioning by adaptive sampling}

\begin{icmlauthorlist}
\icmlauthor{David H. Brookes}{}
\icmlauthor{Hahnbeom Park}{}
\icmlauthor{Jennifer Listgarten}{}
\end{icmlauthorlist}

\thispagestyle{empty}
\section{Algorithm}
\begin{algorithm}[ht]
\caption{{\bf Maximization of a single, continuous property.} 
\mbox{$h_\text{oracle}(\xbf_i)$} is a function returning the expected value of the property oracle.
\mbox{$\text{CDF}_\text{oracle}(\xbf, \gamma$)} is a function to compute the CDF of the oracle predictive model for threshold $\gamma$. $\text{GenProb}(\textbf{x}_i, \textbf{z}_i, \boldtheta)$ is a method that evaluates the probability of an input in a generative model with parameters $\boldtheta$.
$\text{GenTrain}(\{(\xbf_i, w_i)\}$ is a procedure to take weighted training data $\{(\xbf_i, w_i)\}$, and return the parameters of the trained model. $Q$ is a parameter that determines the next iteration's relaxation threshold; $M$ is the number of samples to generate at each iteration. See main text for convergence criteria. $\initset$ is an initial set of samples with which to train the prior. Any line with an $i$ subscript implicitly denotes $\forall i \in [1 \ldots M]$. 
}
\label{ourmethod}
\begin{algorithmic}
\Procedure{$\ourmethod$}{$h_\text{oracle}(\xbf)$, CDF$_\text{oracle}(\xbf, \gamma)$, $\text{GenTrain}(\{\xbf_i\, w_i\})$, $\text{GenProb}(\xbf_i)$, [$Q=0.9$], [$M=100$]} 
\State $\boldthetazero \gets \text{GenTrain}(\initset, {w_i = 1})$ 
\State $\boldphi^{(1)} \gets \boldthetazero$
\State $t \gets 1$
\While {\text{not converged}}
    \State $set_i \gets \xbf_i, \zbf_i \sim G(\boldphit)$
    \State $scores_i \gets h_\text{oracle}(\xbf_i)$
    \State $q \gets \text{index of } Q^\text{th} \text{ percentile of } scores$
    \State $\gamma^{(t)} \gets scores_\text{q}$
    \State \mbox{$weights_i \gets \dfrac{\text{GenProb}(set_i, \boldthetazero)}{\text{GenProb}(set_i, \boldphit)}$}
    \State $weights_i \gets weights_i * (1-CDF_\text{oracle}(\xbf_i, \gamma^{(t)})$)
    \State $\boldphit = \gets \text{GenTrain(} set, weights)$
    \State $t \gets t+1$
\EndWhile
\Return $set, weights$
\cut{
\BState \emph{top}:
\If {$i > \textit{stringlen}$} \Return false
\EndIf
\State $j \gets \textit{patlen}$
\BState \emph{loop}:
\If {$\textit{string}(i) = \textit{path}(j)$}
\State $j \gets j-1$.
\State $i \gets i-1$.
\State \textbf{goto} \emph{loop}.
\State \textbf{close};
\EndIf
\State $i \gets i+\max(\textit{delta}_1(\textit{string}(i)),\textit{delta}_2(j))$.
\State \textbf{goto} \emph{top}.
}
\EndProcedure
\end{algorithmic}
\end{algorithm}

\section{Generalization to specification}

 The \ourmethod~ procedure can be easily extended to perform specification of a property value rather than maximization. In this case the target set is an infinitesimally small range around the target, \ie $S = [y_0 - \epsilon, y_0 + \epsilon]$ for a target value $y_0$ and a small $\epsilon > 0$.  The \ourmethod~procedure remains mostly identical to that of maximization case, except in this case the intermediate sets $S^{(t)} = [y_0 - \gamma^{(t)}, y_0 + \gamma^{(t)}]$ are centered on the specified value and have an update-able width, $\gamma^{(t)}$. The $\gamma^{(t)}$ values are then updated analogously to the thresholds in the maximization case, \ie $\gamma^{(t)}$ is set to the $Q^\text{th}$ percentile of $|y_i - y_0|$ values, for $i=1,...,M$, where $y_i$ are the expected property values according to the sample $\textbf{x}_i \sim p(\textbf{x}|\boldthetat)$.
 
\section{Generalization to multiple properties}

Additionally, \ourmethod~ can be extended to handle multiple properties $Y_1,...,Y_K$ with corresponding desired sets $S_1,...,S_K$. We only require that these properties are conditionally independent given a realization of $X$. In this case, 
\begin{equation}
    P(S_1,...,S_K|\textbf{x}) = \prod_{i=1}^K P(S_i|\textbf{x})
\end{equation}
where now each $Y_i$ has an independent oracle. This distribution, and the corresponding marginal distribution $P(S_1,...,S_K|\boldtheta) = \int d\textbf{x}\, p(\textbf{x}|\boldtheta) \prod_{i=1}^K P(S_i|\textbf{x})$ can then be used in place of $P(S|\xbf)$ and $P(S|\boldtheta)$ in Equations (1)-(13) in the main text to recover the \ourmethod~procedure for multiple properties.

\section{Extension to models not permitting MLE}

Many models cannot be fit with maximum likelihood estimation, and in this case we cannot solve the \ourmethod~ update equation, \eqref{eq:mc_obj}, exactly. However, \ourmethod~ can still be used in the case when approximate inference procedures can be performed on these models, for example any model that can be fit with variational inference \nocite{Jordan1999}\citesupp{Jordan1999}. We derive the \ourmethod~ update equation in the variational inference case below, but the update equation can be modified in a corresponding way for any model that permit other forms of approximate MLE.

In variational inference specifically, the maximum likelihood is lower bounded by an alternative objective:
\begin{align}
    &\max_{\boldphi} \log q(\xbf|\boldphi)
    \\= &\max_{\boldphi} \log \mathbb{E}_{q(\textbf{z})}[q(\xbf|\textbf{z}, \boldphi)] \\ 
    \geq &\max_{\boldphi \boldsymbol{\psi}} \mathbb{E}_{r(\textbf{z}|\xbf, \boldsymbol{\psi})}\left[\log q(\xbf|\textbf{z}, \boldphi)\right] - D_{KL}\left[r(\textbf{z}|\xbf, \boldsymbol{\psi})||q(\textbf{z})\right]  \\
    = &\max_{\boldphi, \boldsymbol{\psi}} \mathcal{L}(\textbf{x}, \boldphi, \boldsymbol{\psi})\label{eq:elbobound}
\end{align}
where $\textbf{z}$ is a realization of a latent variable with prior $p(\textbf{z})$, and $r(\textbf{z}|\xbf, \boldsymbol{\psi})$ is an approximate posterior with parameters $\boldsymbol{\psi}$.  Equation \eqref{eq:elbobound} implies that we can lower bound the the argument of \eqref{eq:mc_obj}:
\begin{equation}\label{eq:gencem_update3}
    \max_{\boldphi} \sum_{i=1}^M \dfrac{p(\xbf^{(t)}_i|\boldthetazero)}{q(\xbf^{(t)}_i|\boldphit)}P(S^{(t)}|\xbf_i^{(t)})\log q(\xbf^{(t)}_i|\boldphi)  \geq \max_{\boldphi, \boldsymbol{\psi}} \sum_{i=1}^M \dfrac{p(\xbf^{(t)}_i|\boldthetazero)}{q(\xbf^{(t)}_i|\boldphit)}P(S^{(t)}|\xbf_i^{(t)}) \mathcal{L}(\textbf{x}_i^{(t)}, \boldphi, \boldsymbol{\psi})
\end{equation}
which is a tight bound when the approximate posterior in the model is rich enough for the approximate posterior to exactly match the true model posterior. This suggests a new update equation, specific for models trained with variational inference:
\begin{equation}
    \boldphi^{(t+1)}, \boldsymbol{\psi}^{(t+1)} = \argmax_{\boldphi, \boldsymbol{\psi}} \sum_{i=1}^M \dfrac{p(\xbf^{(t)}_i|\boldthetazero)}{q(\xbf^{(t)}_i|\boldphit)}P(S^{(t)}|\xbf_i^{(t)}) \mathcal{L}(\textbf{x}_i^{(t)}; \boldphi, \boldsymbol{\psi})
\end{equation}
where we now give time dependence to the approximate posterior parameters, $\boldsymbol{\psi}$. In practice, this is the update equation we use for \ourmethod~ variants that use VAEs as the search distribution (appropriately augmented for latent variables, as described in Section \ref{methods} of the main text). This same process can be used to derive approximate update equations for any weighted maximum likelihood method.

\section{Using Samples from Previous Iterations}

When using model-based optimization methods that update the model parameters using weighted maximum likelihood updates at each iteration, one can extend such a method to make use of samples generated in previous iterations of the algorithm, in the current iteration, so as to potentially increase the effective sample size for the maximum likelihood problem at each iteration. Examples of such methods include that presented herein, \ourmethod, and also \dbas, \rwr~ and \cempi~ (see Section \ref{method_details}). To achieve this goal of using samples from previous iterations, we use an importance sampling scheme to re-scale the weights, $w(\xbf)$,  by the ratio of likelihoods of the sample under the current search model ($t^{\text{th}}$ iteration), to that of the search model from the earlier iteration ($s^{\text{th}}$ iteration). This can be seen as follows:
\begin{align}
    & \mathbb{E}_{p(\xbf|\boldthetat)}[w(\xbf)\log p(\xbf|\boldtheta)] \\
    = &\dfrac{1}{t}\sum_{s=1}^{t} \mathbb{E}_{p(\xbf|\boldthetat)}[w(\xbf)\log p(\xbf|\boldtheta)] \label{eq:keep1}\\
    = &\dfrac{1}{t}\sum_{s=1}^{t} \mathbb{E}_{p(\xbf|\boldtheta^{(s)})}\left[\dfrac{p(\xbf|\boldthetat)}{p(\xbf|\boldtheta^{(s)})}w(\xbf)\log p(\xbf|\boldtheta)\right], \label{eq:keep2}
\end{align}

where \eqref{eq:keep1} uses a bookkeeping trick of averaging identical quantities in order to introduce a sum, and \eqref{eq:keep2} introduces $p(\xbf|\boldtheta^{(s)})$ as an importance sampling proposal density into this sum. Using samples drawn from the model at each iteration, $s$, (\ie old samples) we arrive at the final objective argument sums over the current samples, and samples from all previous iterations:
\begin{align}
    \dfrac{1}{tM} \sum_{s=1}^t \sum_{i=1}^M \dfrac{p(\xbf^{(s)}_i|\boldthetat)}{p(\xbf^{(s)}_i|\boldtheta^{(s)})}w(\xbf^{(s)}_i)\log p(\xbf^{(s)}_i|\boldtheta), \label{eq:keepmc}
\end{align}
where $\xbf^{(s)}_i \sim p(\xbf|\boldtheta^{(s)})$ are $M$ samples drawn at iteration $s$. Note that the same methods outlined in the `Extension to intractable latent variable models' section of the main text can be used to calculate the likelihood ratio, $\frac{p(\xbf^{(s)}_i|\boldthetat)}{p(\xbf^{(s)}_i|\boldtheta^{(s)})}$,  in \eqref{eq:keepmc}, if the marginal likelihoods cannot be calculated exactly (\eg when an Evidence Lower Bound is used such as in the VAE).

Note that as the difference between $t$ and $s$ increases (\ie we start to consider older and older samples), it may be the case that the likelihood ratios, $\dfrac{p(\xbf^{(s)}_i|\boldthetat)}{p(\xbf^{(s)}_i|\boldtheta^{(s)})}$, in \eqref{eq:keepmc} become extremely small. In such cases,  the effective sample size (\ie $\sum_s \sum_i \frac{p(\xbf^{(s)}_i|\boldthetat)}{p(\xbf^{(s)}_i|\boldtheta^{(s)})}w(\xbf^{(s)}_i)$) of the MC estimate \eqref{eq:keepmc} may not be much more than that from an iteration that does not use older samples, $\sum_{i=1}^{M} w(\xbf^{(s)}_i)$. This potentially limits the utility of keeping old samples for many weighting schemes, such as in \dbas, \rwr~ and \cempi.  However, in \ourmethod, where $w(\xbf) = \frac{p(\xbf|\boldthetazero)}{p(\xbf|\boldthetat)}P(S^{(t)}|\xbf)$ at iteration $t$, \eqref{eq:keepmc} becomes:
\begin{align}
    &\dfrac{1}{tM} \sum_{s=1}^t \sum_{i=1}^M \dfrac{p(\xbf^{(s)}_i|\boldthetat)}{p(\xbf^{(s)}_i|\boldtheta^{(s)})}\dfrac{p(\xbf^{(s)}_i|\boldthetazero)}{p(\xbf^{(s)}_i|\boldthetat)}P(S^{(t)}|\xbf^{(s)}_i)\log p(\xbf^{(s)}_i|\boldtheta)  \\
    = & \dfrac{1}{tM} \sum_{s=1}^t \sum_{i=1}^M \dfrac{p(\xbf^{(s)}_i|\boldthetazero)}{p(\xbf^{(s)}_i|\boldtheta^{(s)})}P(S^{(t)}|\xbf^{(s)}_i)\log p(\xbf^{(s)}_i|\boldtheta).
\end{align}
In this case, the likelihood ratio uses the prior rather than the current search model density as in \eqref{eq:keepmc}. Consequently, the old samples get used in the same way as they would have at their own iteration, and their impact does not diminish as iterations proceed, other than by virtue of the factor $P(S^{(t)}|\xbf^{(s)}_i)$. In particular, the only change in the weight of a sample as it gets used in later iterations arises from the updating of the relaxed desired property, $S^{(t)}$. This suggests that it may be more useful to keep old samples in \ourmethod~ than in other methods that update with a weighted ML objective.

It's interesting to note that this correct use of previous samples, by virtue of an importance sampling weight, is conceptually similar to some off-policy procedures in reinforcement learning  \nocite{Precup2001OffPolicyTD, peshkin2002, Tang2010} \citesupp{Precup2001OffPolicyTD, peshkin2002, Tang2010}. 

\section{Experimental Details}

 Here we provide the necessary details to run the experiments described in the main text. In what follows, when we specify model architectures we use the notation \texttt{LayerType(OutputShape)} to describe layers, and the notation $\texttt{Layer1(Out1)} \rightarrow \texttt{Layers2(Out2)}$ to denote that $\texttt{Out1}$ is given as the input to $\texttt{Layer2}$.

\subsection{Methods details}\label{method_details}

\paragraph{Weighted Maximum Likelihood methods}

Similar to \ourmethod~ the methods \rwr, \dbas, and \cempi~ are weighted maximum likelihood methods that update the parameters of a generative model by taking samples from the model, $\xbf_i\sim q(x|\boldphit)$ and optimizing the objective:

\begin{equation}
    \boldphi^{(t+1)} = \argmax_{\boldphi} \sum_{i=1}^M w(\textbf{x}_i)\log q(\xbf_i^{(t)}|\boldphi).
\end{equation}

The methods differ in the definition of these weights:
\begin{itemize}
    \item In \ourmethod, $w(\textbf{x}_i) = \dfrac{p(\xbf|\boldthetazero)}{p(\textbf{x}|\boldthetat)} P(S^{(t)}|\textbf{x}_i)$
    \item In \dbas, $w(\textbf{x}_i) = P(S^{(t)}|\textbf{x}_i)$
    \item In \rwr,  $w(\xbf_i) =  \dfrac{e^{\alpha \mathbb{E}_{p(y|\xbf_i)}[y]}}{\sum_{i=1}^M e^{\alpha \mathbb{E}_{p(y|\xbf_i)}[y]}}$, where $\alpha = 50$ for our experiments (chosen based on a grid search to best optimize the oracle expectation).
    \item In \cempi, $w(\xbf_i) = \mathds{1}_{\{PI(X) \geq \beta^{(t)}\}}(\textbf{x}_i)$ where $PI(\textbf{x})$ is the probability of improvement function and $\beta^{(t)}$ is adjusted according to the methods of CEM (there is a parameter in CEM that corresponds to our Q, which we set to 0.8 for \cempi).
\end{itemize}

 \paragraph{VAE architecture} The following VAE architecture was used for all experiments. The VAE encoder architecture is $\texttt{Input(L, 20)} \rightarrow \texttt{Flatten(L*20)} \rightarrow \texttt{Dense(50)} \rightarrow \texttt{Dense(40)}$. The final output is split into two vectors of length 20, which represent the mean and log-variance of the latent variable, respectively. The decoder architecture is $\texttt{Input(20)} \rightarrow \texttt{Dense(50)} \rightarrow \texttt{Dense(L*20)} \rightarrow \texttt{Reshape(L, 20)} \rightarrow \texttt{ColumnSoftmax(L, 20)}$. Note that the 20 comes from both the number of amino acids, which encode proteins, and the fact that we set the dimensionality of our latent space to be 20.
 
 For the Gomez-Bombarelli methods, this is augmented with a predictive network from the latent space to the property space $\texttt{Input(20)} \rightarrow \texttt{Dense(50)} \rightarrow \texttt{Dense(1)}$

 \paragraph{\fbvae~ parameter settings} A major implementation choice in 
\fbvae~ is the value of the threshold used to 
decide whether to give 0/1 weights to samples. We found that setting the threshold to the 
$80^{\text{th}}$ percentile of the property values in the initial training set gave the best 
performance, and used that setting for all tests presented here.

\paragraph{\fbvae~ implementation} We note that a minor modification to the {\it FB-GAN} framework was required to accommodate a VAE generator instead of a GAN. Specifically we must have the method sample from the distribution 
output by the VAE decoder in order to get sequence realizations, rather than taking the 
argmax of the Gumbel-Softmax approximation output by the WGAN.

 \paragraph{Oracle Details}  

The oracles for the GFP fluorescence task are  neural network ensembles trained according to the method in \citesupp{Lakshmin2017} (without adversarial examples), where each model has the architecture $\texttt{Input(L, 20)} \rightarrow \texttt{Flatten(L*20)} \rightarrow \texttt{Dense(20)} \rightarrow \texttt{Dense(2)}$.

\section{Fluorescence data set}

These data consisted of 50,000 protein sequences each with fluorescent readout. These data showed a clear bimodal distribution of fluorescence values; we retained only the top 34,256 fluorescent proteins (corresponding to the mode with higher fluorescence values) in order to simplify our simulations., consisting of 50,000 protein sequences each with fluorescent readout. These data showed a clear bimodal distribution of fluorescence values; we retained only the top 34,256 fluorescent proteins (corresponding to the mode with higher fluorescence values) in order to simplify our simulations. Also see Figure \ref{fig:figure_s2}

\cut{
\section{Protein stability analysis}

To estimate protein stability for each sequence, initially, the wild-type protein is relaxed in Cartesian space using the Rosetta FastRelax protocol. Then, the best rotameric side-chain conformations for wild-type sequence and mutation are determined, and is followed by FastRelax in Cartesian space allowing movements in the backbones of a three-residue window around all of each mutated residues and all the sidechains in the protein. This allows more degrees of freemdom to move compared to the original implementation\nocite{Park2016}\citesupp{Park2016}, better accounting for larger structural changes anticipated from more than one mutation in the sequence.

We used this approximate stability estimation method to analyze the sequences chosen by each method (\ie, those with the highest oracle values from the tests described in Section \ref{GFP_section} of the main text). Specifically, we calculated the stabilities for each of the nine runs (3 oracles times 3 randomly-initialized runs for each oracle $=$ 9 total sequences) for each method. The average stabilities found from these 9 runs, reported as the difference in free energy, $\Delta \Delta G$, between the wild type GFP sequence and the tested sequence, are shown in Table \ref{ddg_table}. Negative or small positive values of $\Delta \Delta G$ indicate that the tested sequence is more or only slightly less stable than the wild type GFP, while large positive values indicate that it is substantially less stable, and may not even fold properly. We additionally report the average number of mutations away from the wild type of the tested sequences.
}
\cut{
Methods that effectively make use of a prior tend to produce more stable sequences, which tend to have fewer mutations away from the wild type. This result suggests that a prior can help to constrain the search towards reasonable areas of the search space.
}
\cut{
The main point here is that methods which do not effectively make use of a prior can be taken to non-sensible parts of the space. For example, {\it CEM-PI} and {\it RWR} both have rather low stability  (scores greater than 5.0). 

\begin{table}[h]
\begin{tabular}{|l|r|r|}
\hline
\rowcolor[HTML]{C0C0C0} 
{\color[HTML]{000000} Method} & \multicolumn{1}{l|}{\cellcolor[HTML]{C0C0C0}{\color[HTML]{000000} $\Delta \Delta G$ (kcal/mol)}} & \multicolumn{1}{l|}{\cellcolor[HTML]{C0C0C0}{\color[HTML]{000000} Number of Mutations}} \\ \hline 
AM-VAE  & -1.8060    & 17.0     \\ \hline
DbAS    & 0.0805     & 13.55    \\ \hline
GB      & 1.1970     & 2.0      \\ \hline
CbAS    & 1.5804     & 2.55     \\ \hline
FBVAE   & 1.8571     & 4.33     \\ \hline
GB-NO   & 2.7433     & 8.66     \\ \hline
RWR     & 6.4537     & 21.77    \\ \hline
CEM-PI  & 11.4296    & 28.4      \\ \hline
\end{tabular}
\caption{Results of protein stability analysis. The first column reports the method, the second, the average stability score, and the final column reports the average number of mutations of each sequence from the wild type GFP sequence.}
\label{ddg_table}
\end{table}
}
\clearpage

\begin{figure*}[t]
    \centering
    \begin{subfigure}[t]{0.32\textwidth}
        \centering
        \includegraphics[width=\textwidth]{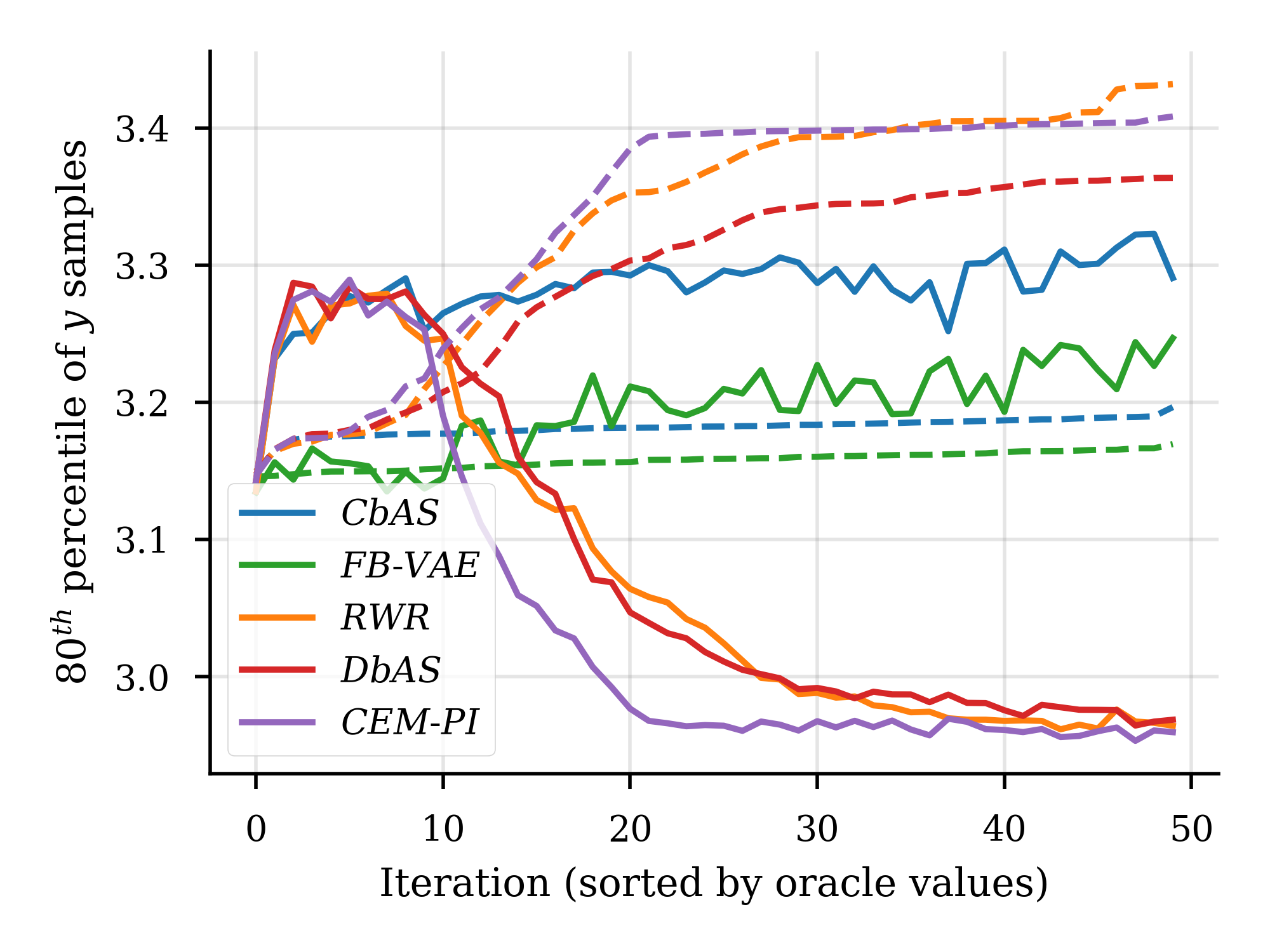}
    \end{subfigure}
    \begin{subfigure}[t]{0.32\textwidth}
        \centering
        \includegraphics[width=\textwidth]{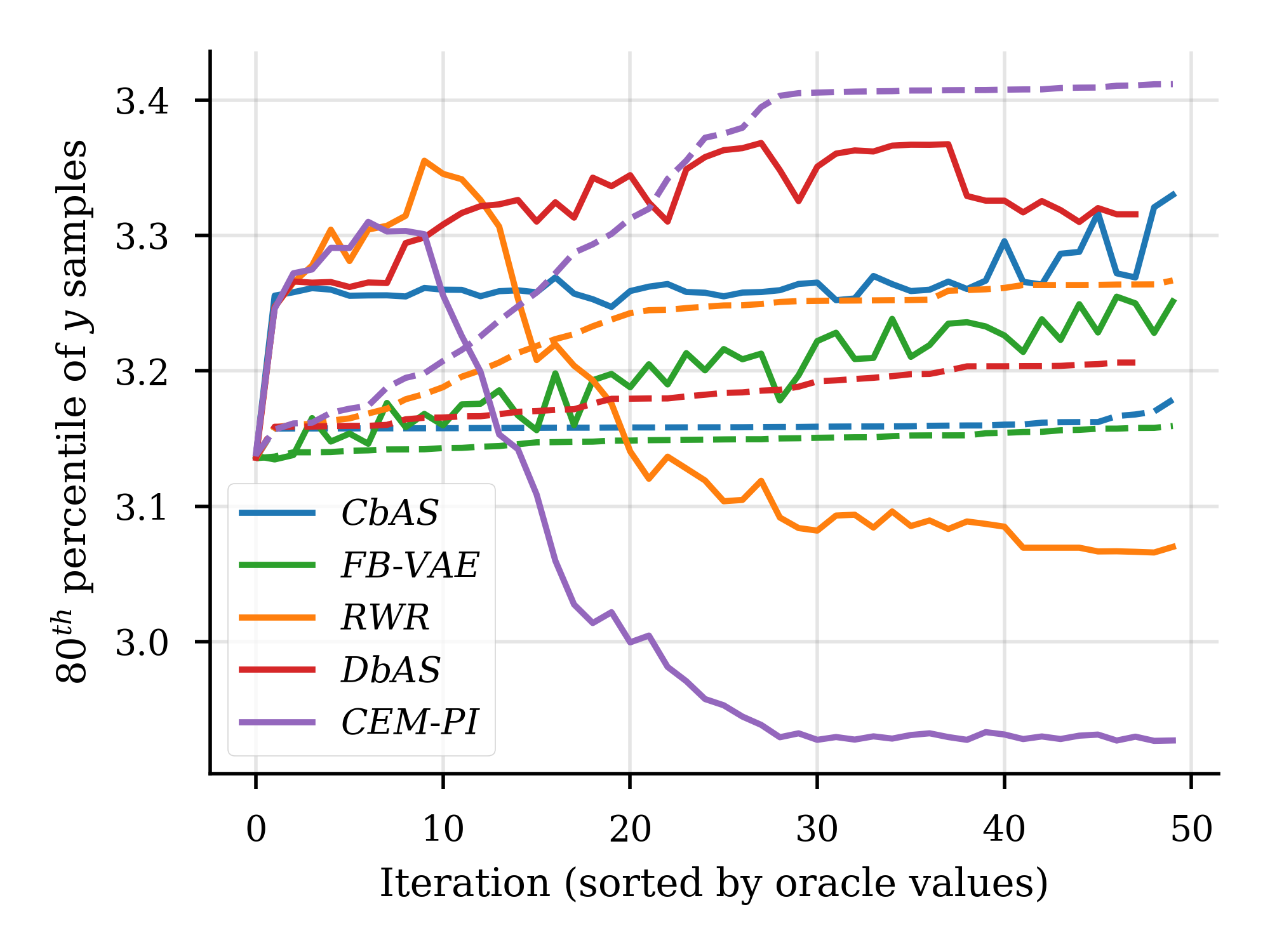}
    \end{subfigure}
    \begin{subfigure}[t]{0.32\textwidth}
        \centering
        \includegraphics[width=\textwidth]{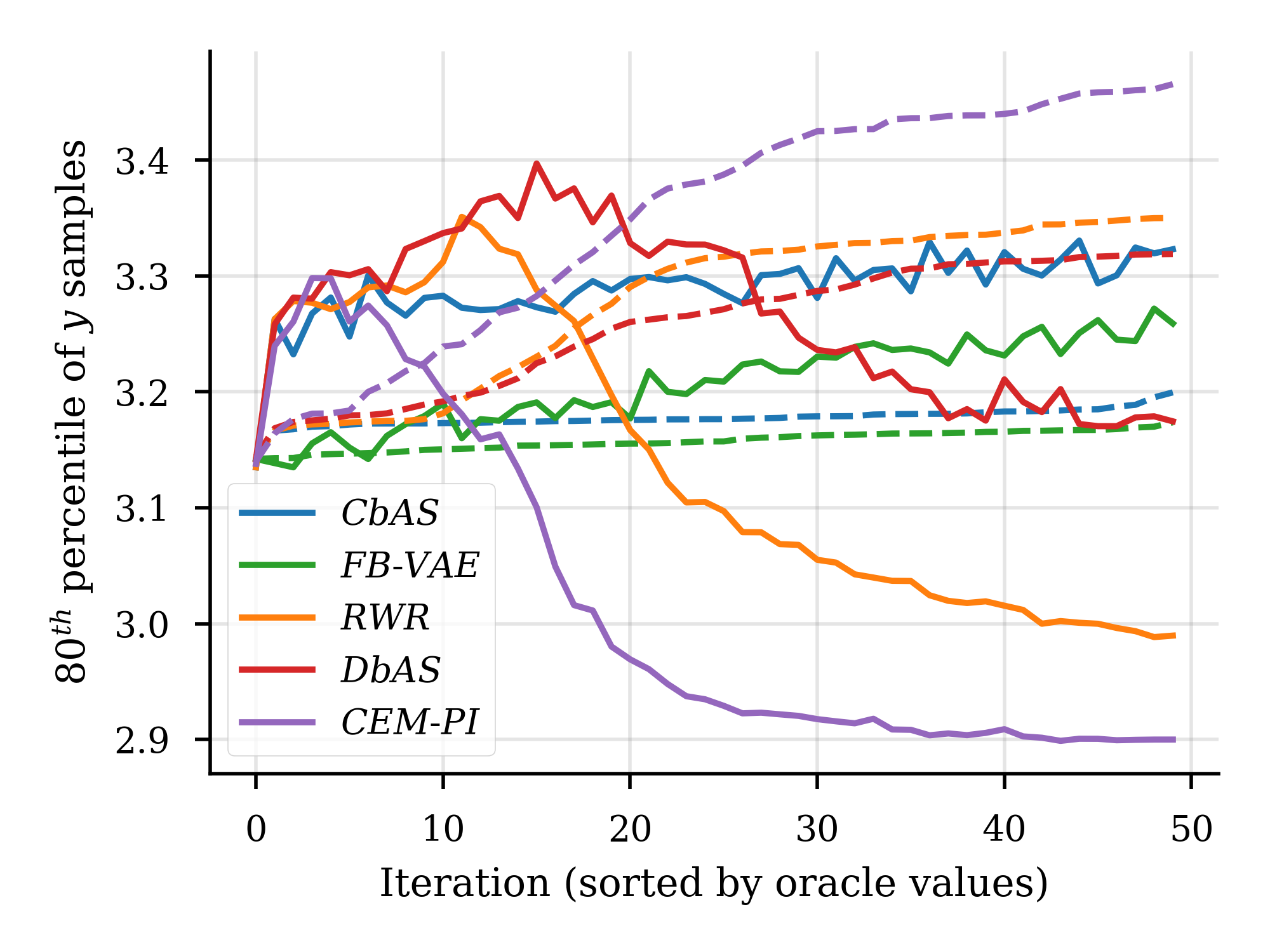}
    \end{subfigure}
    \vskip\baselineskip
    \begin{subfigure}[t]{0.32\textwidth}
        \centering
        \includegraphics[width=\textwidth]{figures/traj_0.8_1_1_all.png}
    \end{subfigure}
    \begin{subfigure}[t]{0.32\textwidth}
        \centering
        \includegraphics[width=\textwidth]{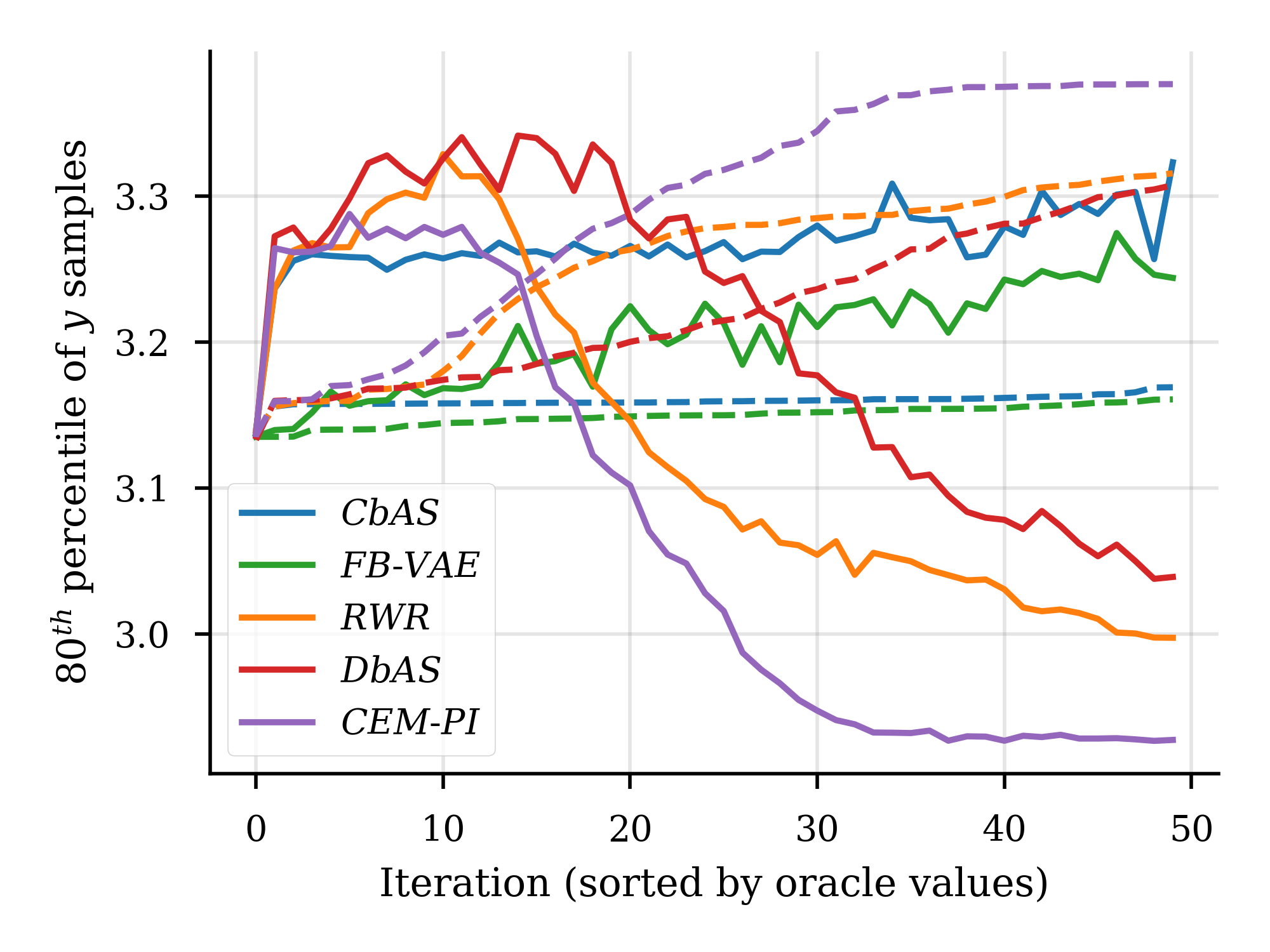}
    \end{subfigure}
    \begin{subfigure}[t]{0.32\textwidth}
        \centering
        \includegraphics[width=\textwidth]{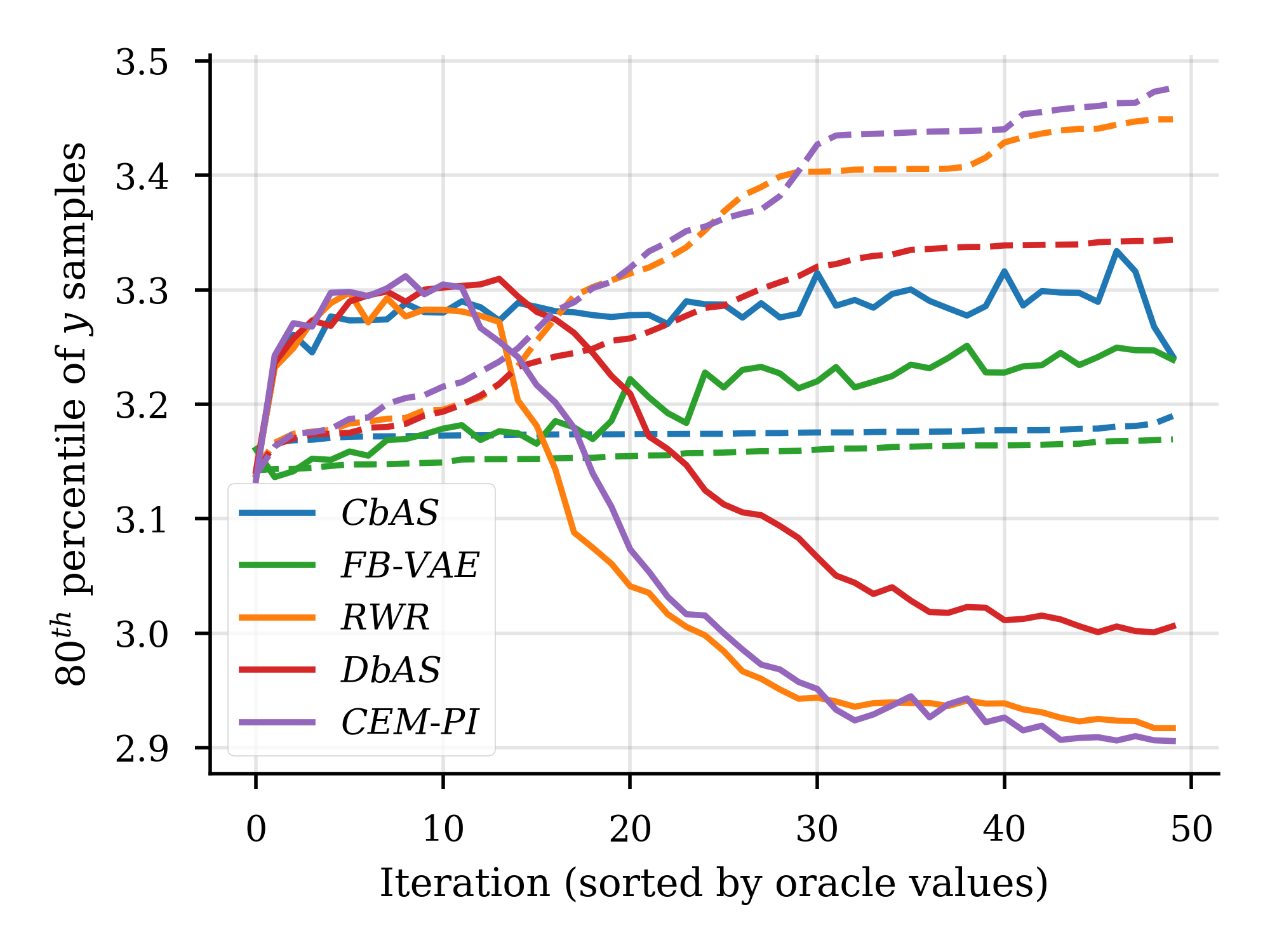}
    \end{subfigure}
    \vskip\baselineskip
    \begin{subfigure}[t]{0.32\textwidth}
        \centering
        \includegraphics[width=\textwidth]{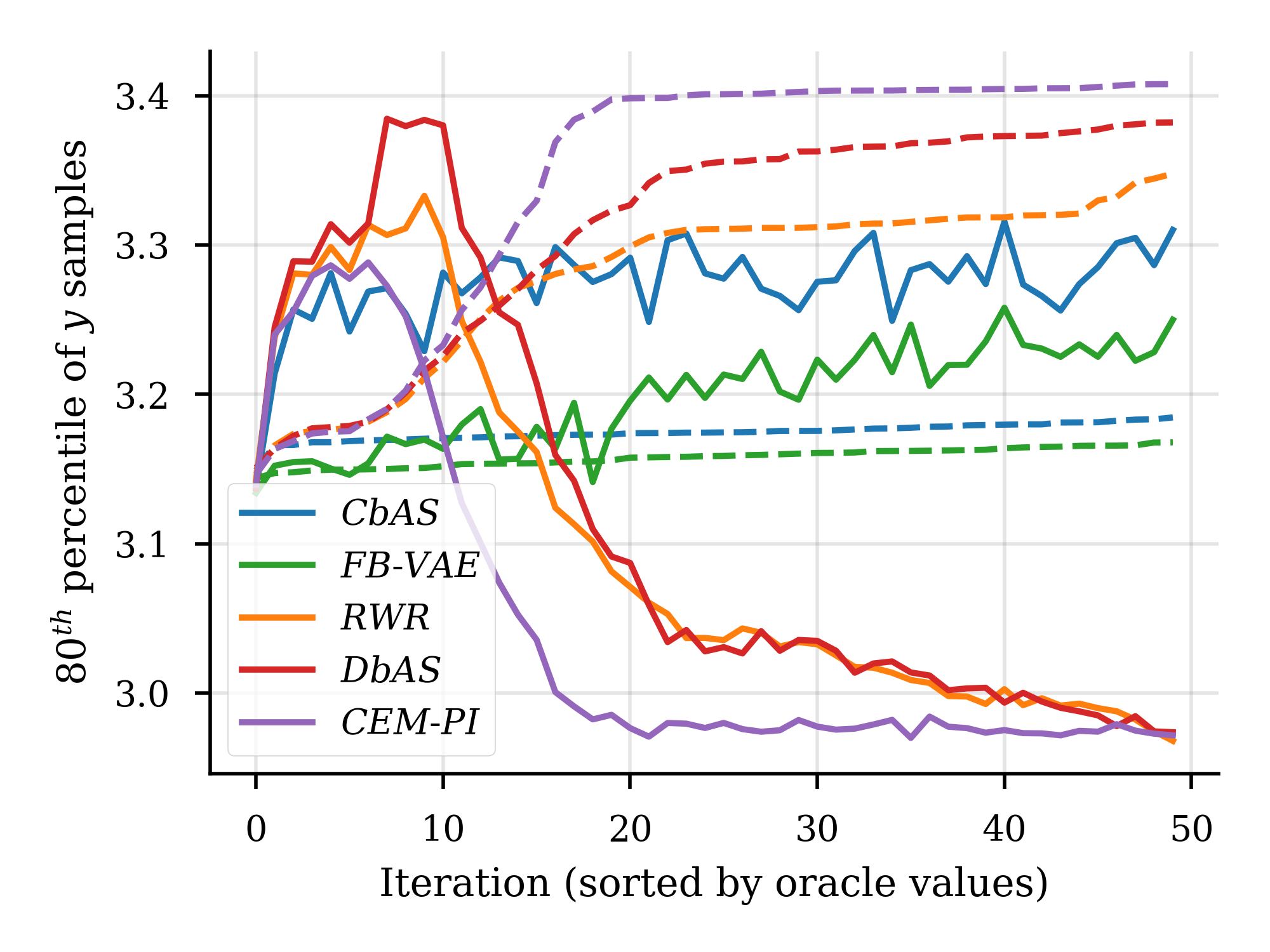}
        \caption{}
    \end{subfigure}
    \begin{subfigure}[t]{0.32\textwidth}
        \centering
        \includegraphics[width=\textwidth]{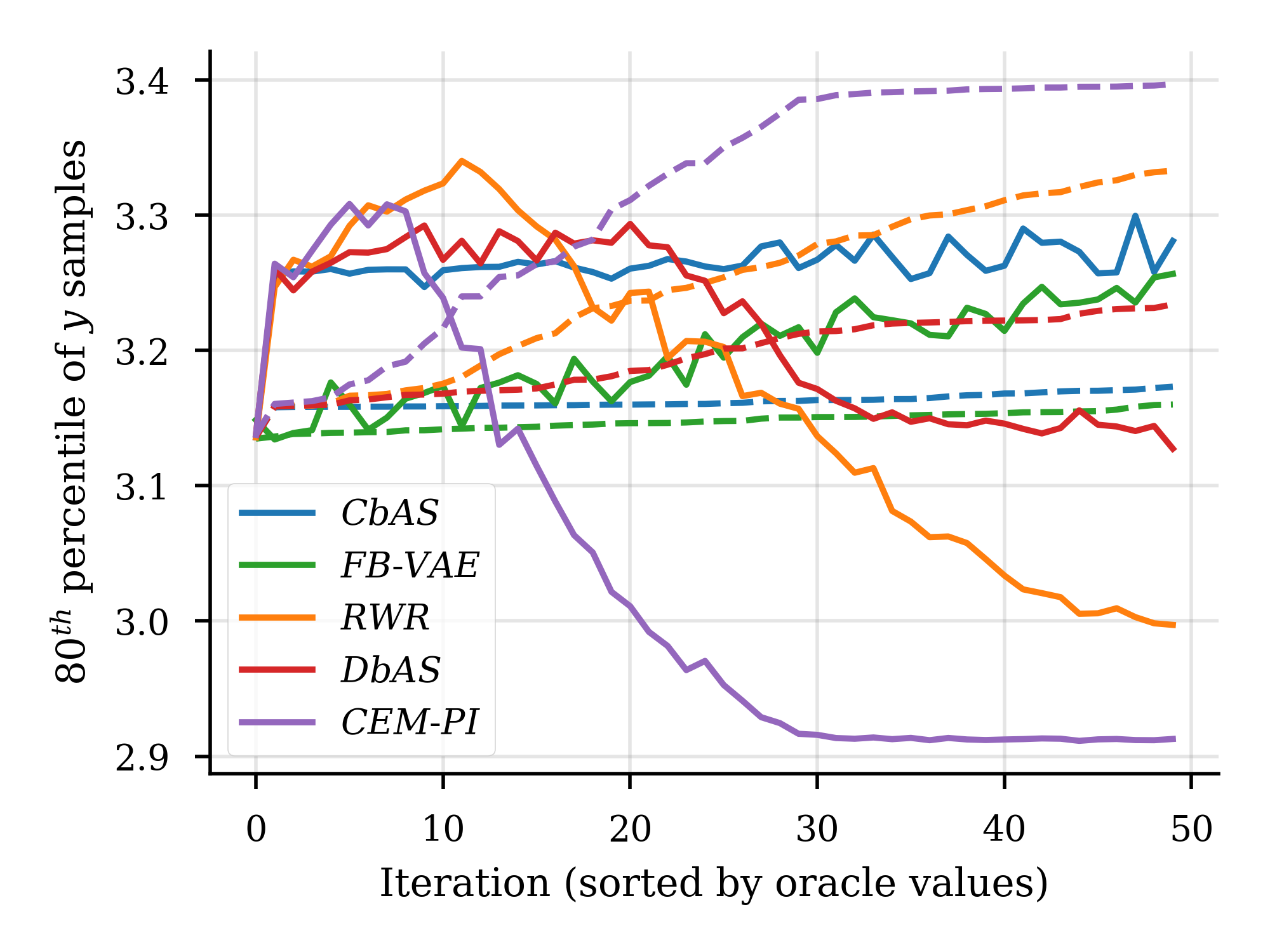}
        \caption{}
    \end{subfigure}
    \begin{subfigure}[t]{0.32\textwidth}
        \centering
        \includegraphics[width=\textwidth]{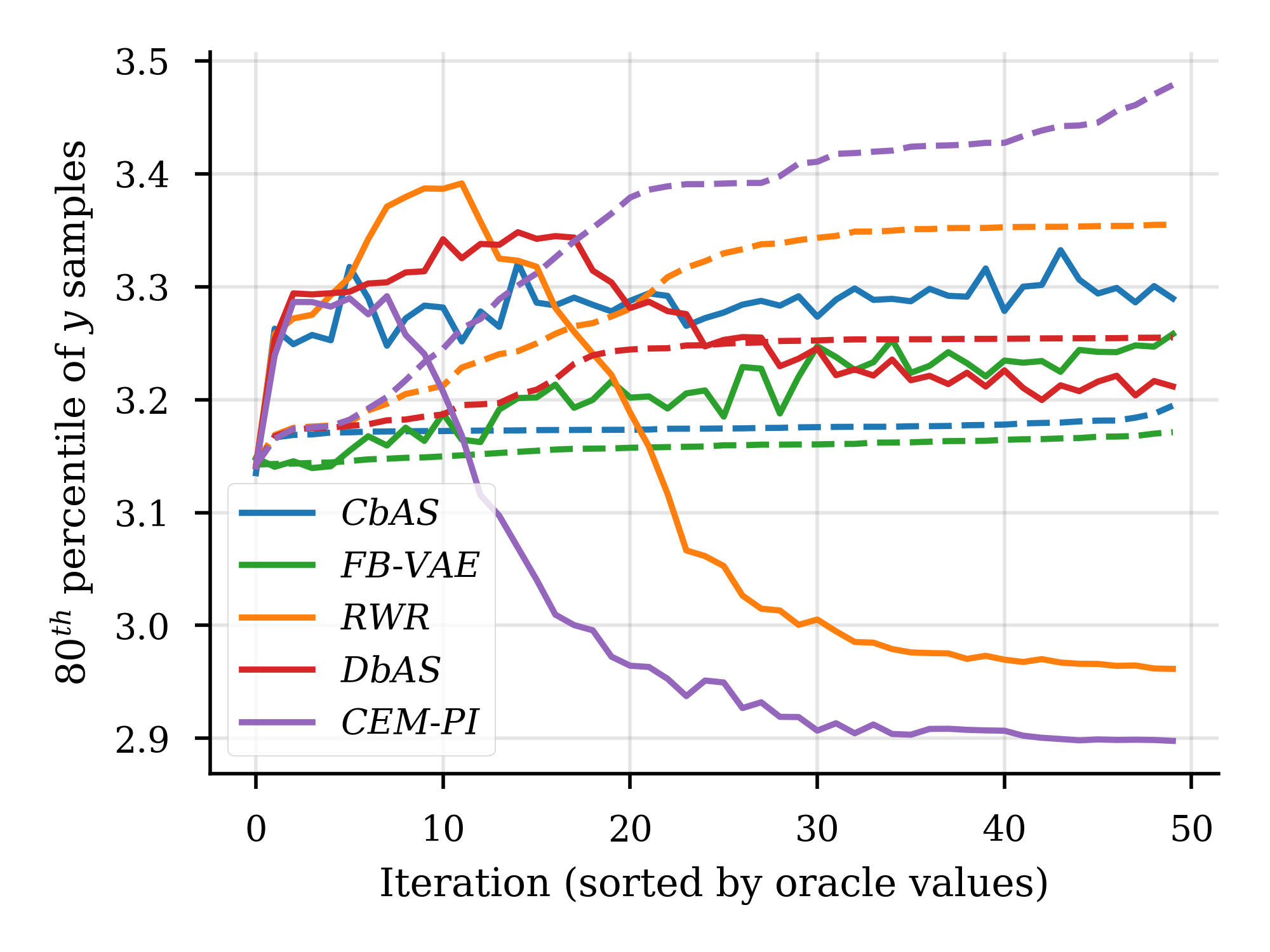}
        \caption{}
    \end{subfigure}
    \caption{Trajectory plots of the same type as Figure \ref{fig:figure2}b (main paper) for all runs of the methods reported in Figure \ref{fig:figure2}a (main paper). Each column shows results for different oracle, namely (a) the ensemble-of-one oracle, (b) the ensemble-of-five oracle, and (c) the ensemble-of-20 oracle. The three columns show three random initializations for each oracle.}
    \label{fig:figure_s1}
\end{figure*}

\clearpage

\begin{figure*}[t]
    \centering
    \begin{subfigure}[t]{0.32\textwidth}
        \centering
        \includegraphics[width=\textwidth]{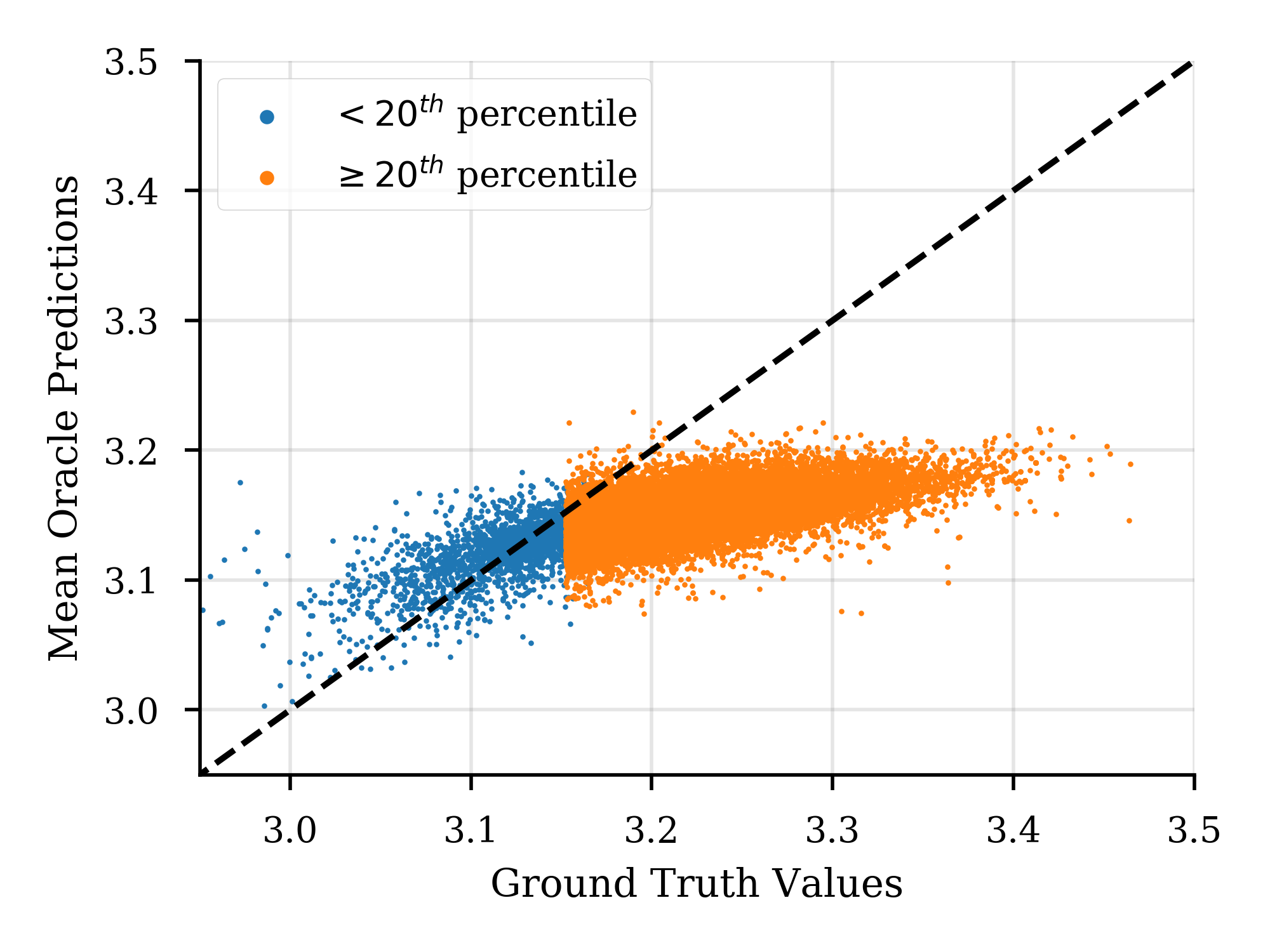}
        \caption{}
    \end{subfigure}
    \begin{subfigure}[t]{0.32\textwidth}
        \centering
        \includegraphics[width=\textwidth]{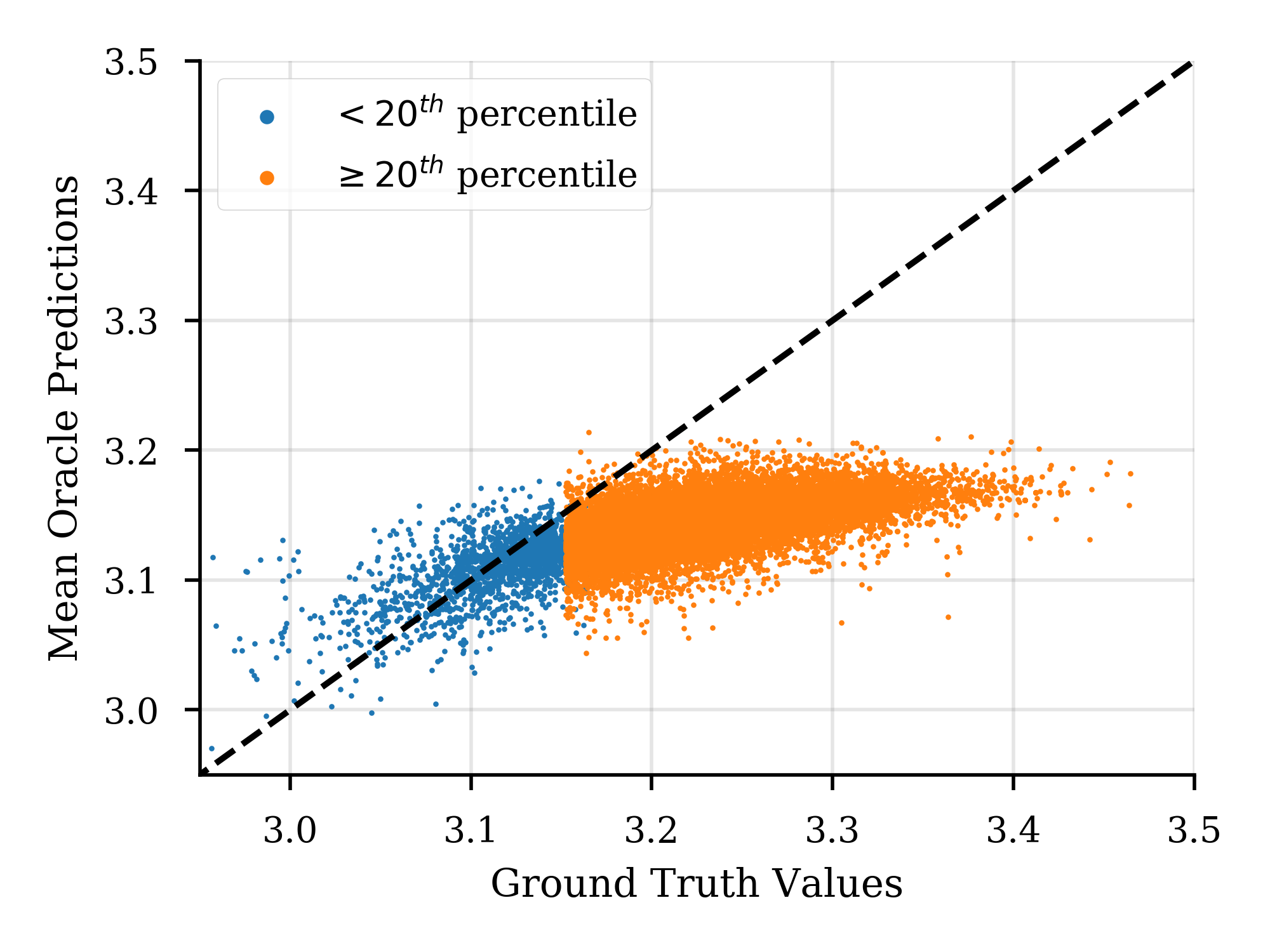}
        \caption{}
    \end{subfigure}
    \begin{subfigure}[t]{0.32\textwidth}
        \centering
        \includegraphics[width=\textwidth]{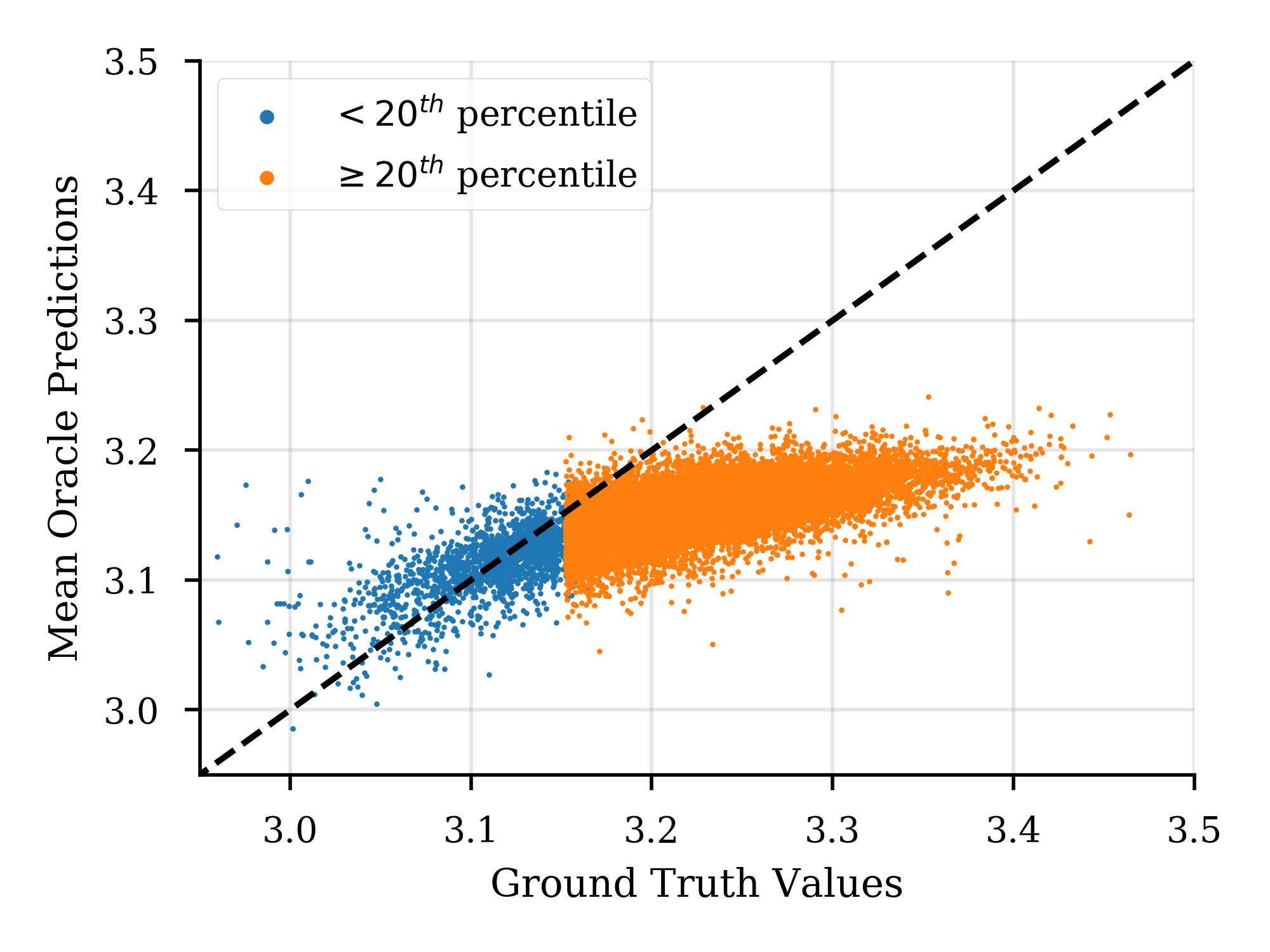}
    \caption{}
    \end{subfigure}
    \caption{Paired plots for each of the three oracles
        described in Section \ref{GFP_section} of the main text. Each point corresponds to a GFP sequence from one of two test sets, described next. The horizontal axis reports the ground truth fluorescence values of the sequence and the vertical axis represents the mean prediction by the oracle for the sequence. 
        Recall that our ground truth data had 6,851 sequences with fluorescence values below the $20^{\text{th}}$ percentile. Five thousand of these were used to train each oracle, and the remaining 1,851 of them are the test dots shown in blue. The orange dots are all 27,404 sequences with fluorescence equal to or above the $20^{\text{th}}$ percentile (oracles were not trained on data in this range). 
        The plots correspond to (a) the ensemble-of-one oracle, (b) the ensemble-of-five oracle, and (c) the ensemble-of-20 oracle. We can see that all oracles are severely biased outside of the training distribution, which is one of the pathologies that \ourmethod~ is designed to avoid.}
    \label{fig:figure_s2}
\end{figure*}

\clearpage

\begin{figure*}[t]
    \centering
    \includegraphics[width=0.75\textwidth]{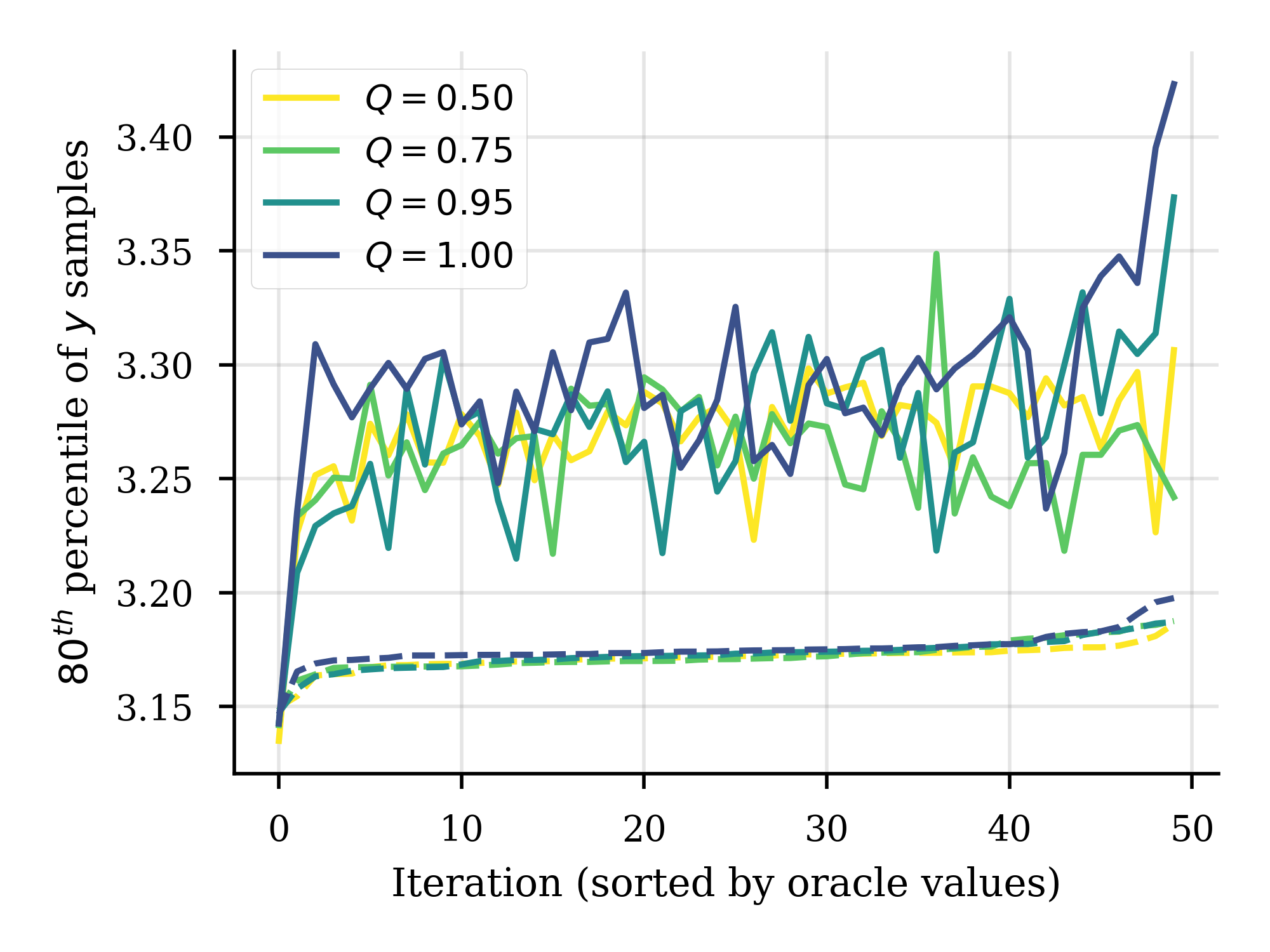}
    \caption{A comparison of \ourmethod~ trajectories for different value of $Q$ (the quantile threshold parameter described in the main text). Each of these trajectories was generated using the ensemble-of-one for the oracle. We see that \ourmethod~ is relatively insensitive to the setting of $Q$; that is, in this example, it would be sufficient to use anything in the range $[0.5,1.0]$.}
    \label{fig:figure_s3}
\end{figure*}

\clearpage

\bibliographysupp{main}
\bibliographystylesupp{icml2019}


\end{document}